\documentclass[final]{cvpr}

\usepackage{times}
\usepackage{epsfig}
\usepackage{graphicx}
\usepackage{amsmath}
\usepackage{amssymb}

\usepackage{subfigure}
\usepackage{pifont}
\usepackage{color, colortbl}
\newcommand{\cmark}{\ding{51}}%
\newcommand{\xmark}{\ding{55}}%

\DeclareMathOperator*{\argmin}{arg\, min}

\usepackage{csquotes}

\usepackage{multirow}
\usepackage{graphicx}
\usepackage{booktabs}

\newcommand*{\twoelementtable}[3][l]%
{%
    \renewcommand{\arraystretch}{0.8}%
    \begin{tabular}[t]{@{}#1@{}}%
        #2\tabularnewline
        #3%
    \end{tabular}%
}
\usepackage{array}
\usepackage{tabularx}

\usepackage{algorithm}
\usepackage{algorithmic}

\usepackage[pagebackref=true,breaklinks=true,colorlinks,bookmarks=false]{hyperref}

\begin{document}

\title{FSDR: Frequency Space Domain Randomization for Domain Generalization}

\author{Jiaxing Huang,\hspace{2mm} Dayan Guan,\hspace{2mm} Aoran Xiao,\hspace{2mm} Shijian Lu\thanks{Corresponding author (Shijian.Lu@ntu.edu.sg).} \vspace{0.5cm} \\ 
School of Computer Science Engineering, Nanyang Technological University\\
}

\maketitle

\begin{abstract}
Domain generalization aims to learn a generalizable model from a `known’ source domain for various `unknown’ target domains. It has been studied widely by domain randomization that transfers source images to different styles in spatial space for learning domain-agnostic features. However, most existing randomization uses GANs that often lack of controls and even alter semantic structures of images undesirably. Inspired by the idea of JPEG that converts spatial images into multiple frequency components (FCs), we propose Frequency Space Domain Randomization (FSDR) that randomizes images in frequency space by keeping domain-invariant FCs (DIFs) and randomizing domain-variant FCs (DVFs) only. FSDR has two unique features: 1) it decomposes images into DIFs and DVFs which allows explicit access and manipulation of them and more controllable randomization; 2) it has minimal effects on semantic structures of images and domain-invariant features. We examined domain variance and invariance property of FCs statistically and designed a network that can identify and fuse DIFs and DVFs dynamically through iterative learning.
Extensive experiments over multiple domain generalizable segmentation tasks show that FSDR achieves superior segmentation and its performance is even on par with domain adaptation methods that access target data in training.
\end{abstract}

\begin{figure}[ht]
\centering
\subfigure {\includegraphics[width=.98\linewidth]{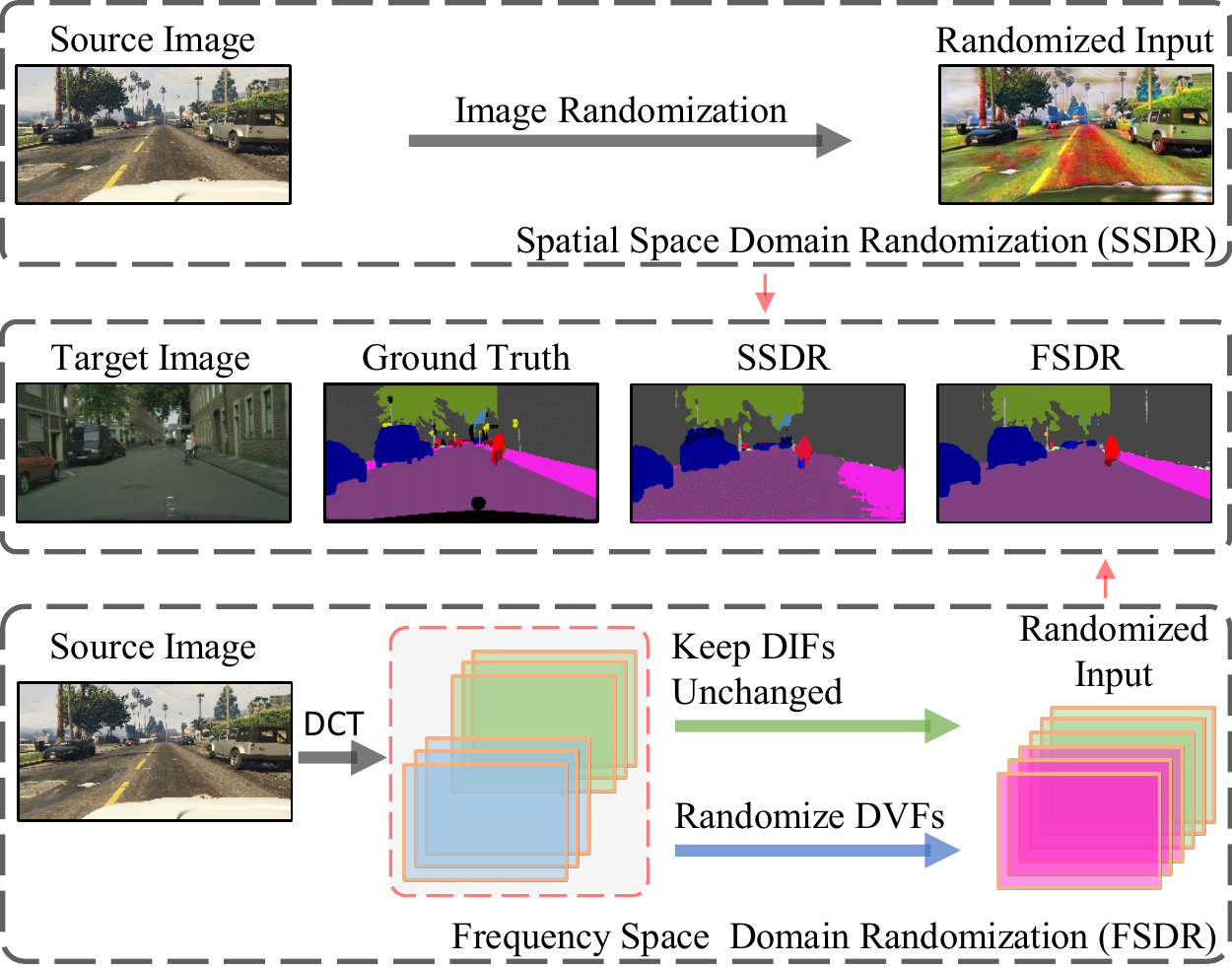}}
\vspace{0.01 pt}
\caption{
Our proposed frequency space domain randomization (FSDR) converts images into multiple frequency components (FCs) with discrete cosine transform (DCT) and identifies domain variant FCs (DVFs) and domain invariant FCs (DIFs). Such explicit isolation allows it to randomize DVFs while keeping DIFs unchanged in training which often leads to more generalizable models. Traditional spatial space domain randomization (SSDR) tends to produce sub-optimal segmentation as it randomizes images as a whole without isolating and preserving domain invariant features. Green, blue and red boxes denotes DIFs, DVFs and randomized DVFs, respectively (best viewed in color).
}
\label{fig:intro}
\end{figure}

\section{Introduction}
Semantic segmentation has been a longstanding challenge in computer vision research, which aims to assign a class label to each and every pixel of an image. Deep learning based methods \cite{chen2017deeplab,long2015fully} have achieved great successes at the price of large-scale densely-annotated training data \cite{cordts2016cityscapes} that are usually prohibitively expensive and time-consuming to collect. One way of circumventing this constraint is to employ synthetic images with automatically generated labels \cite{ros2016synthia,richter2016playing} in network training. However, such models usually undergo a drastic performance drop while applied to real-world images \cite{zhang2016understanding} due to the domain bias and shift \cite{saito2018maximum,luo2018macro, tsai2018learningb, vu2018memory, Saito_2018_ECCV}.

Unsupervised domain adaptation (UDA) has been studied widely to tackle the domain mismatch problem by learning domain invariant/aligned features from a labelled source domain and an unlabelled target domain~\cite{hoffman2016fcns,kang2019contrastive,kang2018deep,sankaranarayanan2018learning,tsai2018learning,tzeng2017adversarial, luo2019taking, vu2019advent, Chen_2018_CVPR,chen2018domain}. However, its training requires target-domain data which could be hard to collect during the training stage such as autonomous driving at various cities, robot exploration of various new environments, etc. Additionally, it is not scalable as it requires network re-training or fine-tuning for each new target domain. Domain generalization has attracted increasing attention as it learns domain invariant features without requiring target-domain data in training~\cite{yue2019domain,muandet2013domain,gan2016learning,li2018domain,li2017deeper,li2017learning}. One widely adopted generalization strategy is domain randomization (DR) that learns domain-agnostic features by randomizing or stylizing source-domain images via adversarial perturbation, generative adversarial networks (GANs), etc.~\cite{shankar2018generalizing,volpi2018generalizing,yue2019domain,qiao2020learning}. However, most existing DR methods randomize the whole spectrum of images in the spatial space which tends to modify domain invariant features undesirably.

We propose an innovative frequency space domain randomization (FSDR) technique that transforms images into frequency space and performs domain generalization by identifying and randomizing domain-variant frequency components (DVFs) while keeping domain-invariant frequency components (DIFs) unchanged. FSDR thus overcomes the constraints of most existing domain randomization methods which work over the full spectrum of images in the spatial space and tend to modify domain-invariant features undesirably as illustrated in Fig.~\ref{fig:intro}. We explored two different approaches for domain randomization in the frequency space. The first is spectrum analysis based FSDR (FSDR-SA) that identifies DIFs and DVFs through empirical studies.
The second is spectrum learning based FSDR (FSDR-SL) that identifies DIFs and DVFs through dynamic and iterative learning processes. Extensive experiments show that FSDR improves the model generalization greatly.  Additionally, FSDR is complementary to spatial-space domain generalization and the combination of the two improves model generalization consistently. 

The contributions of this work can be summarized in three aspects. \textit{First}, we propose an innovative frequency space domain randomization technique that transforms images into frequency space and achieves domain randomization by changing DVFs only while keeping DIFs unchanged. \textit{Second}, we design two randomization approaches in the frequency space that identify DVFs and DIFs effectively through empirical experiments and dynamic learning, respectively. \textit{Third}, extensive experiments over multiple domain generalization tasks show that our proposed frequency space domain randomization technique achieves superior semantic segmentation consistently.


\section{Related Works}

\begin{figure*}[t]
\centering
\subfigure {\includegraphics[width=1\linewidth]{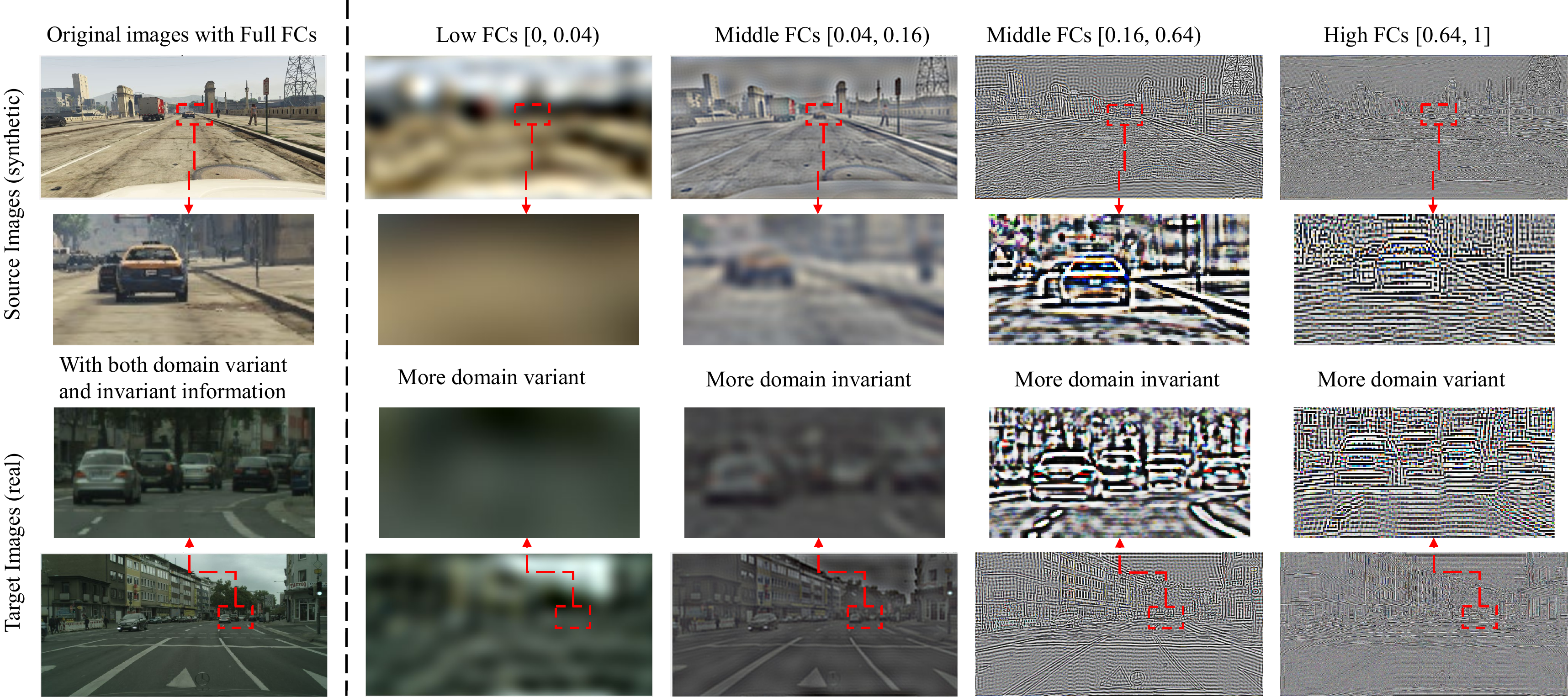}}
\vspace{-10pt}
\caption{
Visualization of spectrum decomposition of synthetic and real images: Column 1 shows \textit{Source Images (synthetic)} and \textit{Target Images (real)} that contains both domain variant and domain invariant information. Columns 2 and 4 show low-pass (\ie, [0, 0.04)) and high-pass (\ie, [0.64, 1]) filtered images that largely capture domain-variant information such as colors and styles. Columns 3 and 4 show middle-pass (\ie, [0.04, 0.16) and [0.16, 0.64)) filtered images that largely capture domain invariant information such as structures and shapes. Images in Rows 2 and 3 are close-up views of highlighted regions in images in Rows 1 and 2.
}
\vspace{-10pt}
\label{fig:visualspectrum}
\end{figure*}

\textbf{Domain Generalization (DG)} aims to learn a generalizable model from ‘known’ source domains for various ‘unknown’target domains \cite{muandet2013domain,gan2016learning}. Most existing DG methods can be broadly categorized into multi-source DG~\cite{muandet2013domain,gong2013reshaping,ghifary2015domain,li2017learning,grubinger2017multi,li2017deeper,mancini2018robust,mancini2018best,shankar2018generalizing,li2018domain,carlucci2019domain,dou2019domain,li2019episodic} and single-source DG~\cite{tobin2017domain,volpi2018generalizing,yue2019domain,qiao2020learning} both of which strive to learn domain-agnostic features by either learning a domain-invariant feature space ~\cite{muandet2013domain,gong2013reshaping,ghifary2015domain,grubinger2017multi,li2017learning,shankar2018generalizing,li2018domain,dou2019domain,carlucci2019domain,zhao2020knowledge} or aggregating domain specific modules ~\cite{li2017deeper,mancini2018robust,mancini2018best,li2019episodic}. Specifically, multi-source DG handles generalization by joint supervised and unsupervised learning \cite{carlucci2019domain}, domain perturbation \cite{shankar2018generalizing}, adversarial feature learning \cite{li2018domain}, meta-learning \cite{li2017learning}, new domain discovery \cite{gong2013reshaping} and network duplication \cite{li2017deeper,li2019episodic}. Single-source DG handles generalization by domain randomization that augment data \cite{tobin2017domain,volpi2018generalizing} or domains \cite{yue2019domain,qiao2020learning}. Our work belongs to single-source DG, aiming to tackle a more challenging scenario in domain generalization when only one single source domain is available in training.

\textbf{Domain Randomization} (DR) is the common strategy in domain generalization \cite{tobin2017domain,shankar2018generalizing,volpi2018generalizing,yue2019domain,qiao2020learning} especially when a single source domain is available \cite{tobin2017domain,volpi2018generalizing,yue2019domain,qiao2020learning}. One early approach achieves domain randomization by manually setting different synthesis parameters for generating various images in certain image simulation environments~\cite{tobin2017domain,sadeghi2016cad2rl,prakash2019structured,sundermeyer2018implicit,tremblay2018training}. The manual approach has limited scalability and the variation of the generated images is constrained as compared with natural images. Another approach is learning based which aims to enrich the variation of synthetic images in a source domain by gradient-based domain perturbation \cite{shankar2018generalizing}, adversarial data augmentation~\cite{volpi2018generalizing}, GANs-based domain augmentation~\cite{yue2019domain} or adversarial domain augmentation~\cite{qiao2020learning}. These methods have better scalability but they randomize the whole spectrum of images in the spatial space which could modify domain invariant features undesirably.
Our method transforms images into frequency space and divides images into DIFs and DVFs explicitly. It randomizes images by modifying DVFs without changing DIFs which has minimal effects over (domain-invariant) semantic structures of images. 

\textbf{Domain Adaptation} is closely relevant to domain generalization but it exploits (unlabelled) target-domain data during the training process. The existing domain adaptation methods can be broadly classified into three categories. The first is \textit{adversarial training} based which employs adversarial learning to align source and target distributions in the feature, output or latent space \cite{hoffman2016fcns,long2016unsupervised,tzeng2017adversarial,luo2019taking,tsai2018learning,chen2018road,zhang2017curriculum,saito2017adversarial,saito2018maximum,vu2019advent,cao2019pedestrian,tsai2019domain,huang2020contextual}. The second is \textit{image translation} based which translates images from source domains to target domains to mitigate domain gaps \cite{hoffman2018cycada, sankaranarayanan2018learning, li2019bidirectional,Zhang2019lipreading,hong2018conditional,yang2020fda}. The third is \textit{self-training} based which employs pseudo label re-training or entropy minimization to guide iterative learning over unlabelled target-domain data \cite{zou2018unsupervised,saleh2018effective,zhong2019invariance,guan2019unsupervised,zou2019confidence,guan2021scale}.



\section{Method}
This section presents our Frequency Space Domain Randomization (FSDR) technique, which consists of three major subsections on spectrum analysis that identifies DIFs and DVFs statically, spectrum analysis in FSDR that studies how DIFs and DVFs work for domain randomization, and spectrum learning in FSDR that shows how domain randomization can be achieved via spectrum learning.

\subsection{Problem Definition}
We focus on the problem of unsupervised domain generalization (UDG) in semantic segmentation. Given source-domain data $X_{s} \subset \mathbb{R}^{H \times W \times 3}$ with C-class pixel-level segmentation labels $Y_{s} \subset (1, C)^{H \times W}$, our goal is to learn a semantic segmentation model $G$ that well performs on unseen target-domain data $X_{t}$. The baseline model is trained with the original source-domain data only: 

\begin{equation}
\mathcal{L}_{orig} = l(G(x_{s}), y_{s}),    
\end{equation}
where $l()$ denotes the standard cross entropy loss.

\subsection{Spectrum Analysis}
This subsection describes spectrum analysis for identifying DIFs and DVFs. For each source-domain image $x_{s} \subset \mathbb{R}^{H \times W \times 3}$, we first convert it into frequency space with Discrete Cosine Transform (DCT) and then decompose the converted signals into 64 FCs $f_{s} \subset \mathbb{R}^{H \times W \times 3 \times 64}$ with a band-pass filter $\mathcal{B}_{p}$:
\begin{equation}
f_{s} = \mathcal{B}_{p}(\mathcal{D}(x_{s})) = \{f_{s}(0), f_{s}(1), f_{s}(3), ..., f_{s}(63)\},
\end{equation}

where $\mathcal{D}$ denotes DCT, $f_{s}(i) \subset \mathbb{R}^{H \times W \times 3}, i = \{0, 1, ..., 63\}$ denote the frequency-wise representation, and the definition of $\mathcal{B}_{p}()$ is in appendix.

\renewcommand\arraystretch{1.1}
\begin{table}[t]
\caption{
We analyze and identify domain variant and invariant frequency components (FCs) by training models with certain FCs on source domain (synthetic), and testing with target domain images (real) in the classification task. The ``source acc." means the test accuracy on the source SYNTHIA dataset while the ``target acc." represents the test accuracy on the target ImageNet dataset.
}
\centering
\begin{tabular}{ccc}
\hline
\hline
\multicolumn{3}{c}{Band-reject Spectrum analysis}
\\\hline
Rejected band & \multicolumn{1}{c}{Source acc.} & \multicolumn{1}{c}{Target acc.}
\\\hline
Null (with full FCs)    & 95.5  &65.2\\\hline
{[}0, 1{)}    & 95.1  &68.6\\
{[}1, 2{)}    & 95.3  &67.1\\
{[}2, 4{)}    & 95.4  &62.3\\
{[}4, 8{)}    & 95.4  &62.7\\
{[}8, 16{)}   & 95.5  &64.6\\
{[}16, 32{)}  & 95.6  &64.9\\
{[}32, 64{]}  & 95.9  &67.4\\
\hline
\end{tabular}
\label{tab:fc_analysis}
\end{table}

We identify DIFs and DVFs in $f_{s}$ by a set of control experiments as shown in Table \ref{tab:fc_analysis}. For each source image, we first filter out FCs with indexes between certain lower/upper threshold (under `Rejected bands’ in Table \ref{tab:fc_analysis}) with a band reject filter $\mathcal{B}_{r}$ and then train models with remaining FCs. The band reject filter $f_{s}^{'} = \mathcal{B}_{r}(f_{s};I)$ can be defined by:
\begin{equation}
f_{s}^{'}(i) =
\left\{
\begin{array}{ll}
0, & \text{if} I(i) = 0,\\
f_{s}(i), & \text{otherwise,}\\
\end{array}
\right.
\end{equation}
where $I$ is a 64-dimensional binary mask vector whose values are 1 for the preserved components and 0 for the discarded.

We then apply the trained model to the target images to examine the domain invariance and generalization of the filtered source-domain FCs. Specifically, improved (or degraded) performance over the target data means that the removed FCs are domain variant (or invariant) and removing them prevents learning domain variant (or invariant) features and improves (or degrades) generalization. With such spectrum analysis experiments with different filter masks $I$, DIFs and DVFs $f = \{f_{var}, f_{invar}\}$ can be identified and recorded by a binary mask vector $I^{SA}$ as follows:
\begin{align}
& x^{f}_{s} = \mathcal{D}^{-1}(\mathcal{B}_{p}(\mathcal{D}(x_{s}))),\nonumber\\
& G^{f} = \argmin_{\theta} l(G(x^{f}_{s};\theta), y_{s}),\nonumber\\
& x^{f'}_{s} = \mathcal{D}^{-1}(\mathcal{B}_{r}(\mathcal{B}_{p}(\mathcal{D}(x_{s}));I)),\nonumber\\
& G^{f'} = \argmin_{\theta} l(G(x^{f'}_{s};\theta), y_{s}),\nonumber\\
& I^{SA}(I) = \left\{
\begin{array}{ll}
0, & \text{if} \ \alpha(G^{f'}(x^{f}_{t}), y_{t}) > \alpha(G^{f}(x^{f}_{t}), y_{t}),\\
1, & \text{if} \ \alpha(G^{f'}(x^{f}_{t}), y_{t}) < \alpha(G^{f}(x^{f}_{t}), y_{t}),\\
\end{array}
\right.\nonumber
\end{align}
where $\alpha()$ denotes a binary function that evaluate the prediction accuracy (it returns $1.0$ for correct prediction and $0.0$ otherwise), $x^{f}_{s}$/$x^{f}_{t}$ is the input with full FCs from the source/target domain, $x^{f'}_{s}$ is the filtered input (\ie, $f_{s}^{'} = \mathcal{B}_{r}(f_{s};I)$) from source domain, $y_{s}/y_{t}$ is the source/target ground truth, and $I^{SA}$ is a 64-dimensional binary vector telling whether a FC is domain invariant (\ie, $I^{SA} = 1$) or variant (\ie, $I^{SA} = 0$). The \textit{Spectrum Analysis} algorithm is available in appendix \textcolor{red}{A.1.}

We evaluated the spectrum analysis over a synthetic-to-real domain generalizable image classification task (\ie, cropped SYNTHIA to ImageNet). Table~\ref{tab:fc_analysis} shows experimental results.

We can observe that removing high-frequency and low-frequency components both improve the model generalization clearly. In addition, we visualize the spectrum decomposition of source (\ie, GTA) and target (\ie, Cityscapes) images in Fig.~\ref{fig:visualspectrum}. We can observe that low-frequency and high-frequency components capture more domain variant features (in between GTA and Cityscapes) as compared with middle-frequency components.
Note several prior works exploited different FC properties successfully in other tasks such as data compression \cite{xu2020learning} and supervised learning \cite{wang2020high}.

\begin{figure*}[t]
\centering
\subfigure {\includegraphics[width=0.99\linewidth]{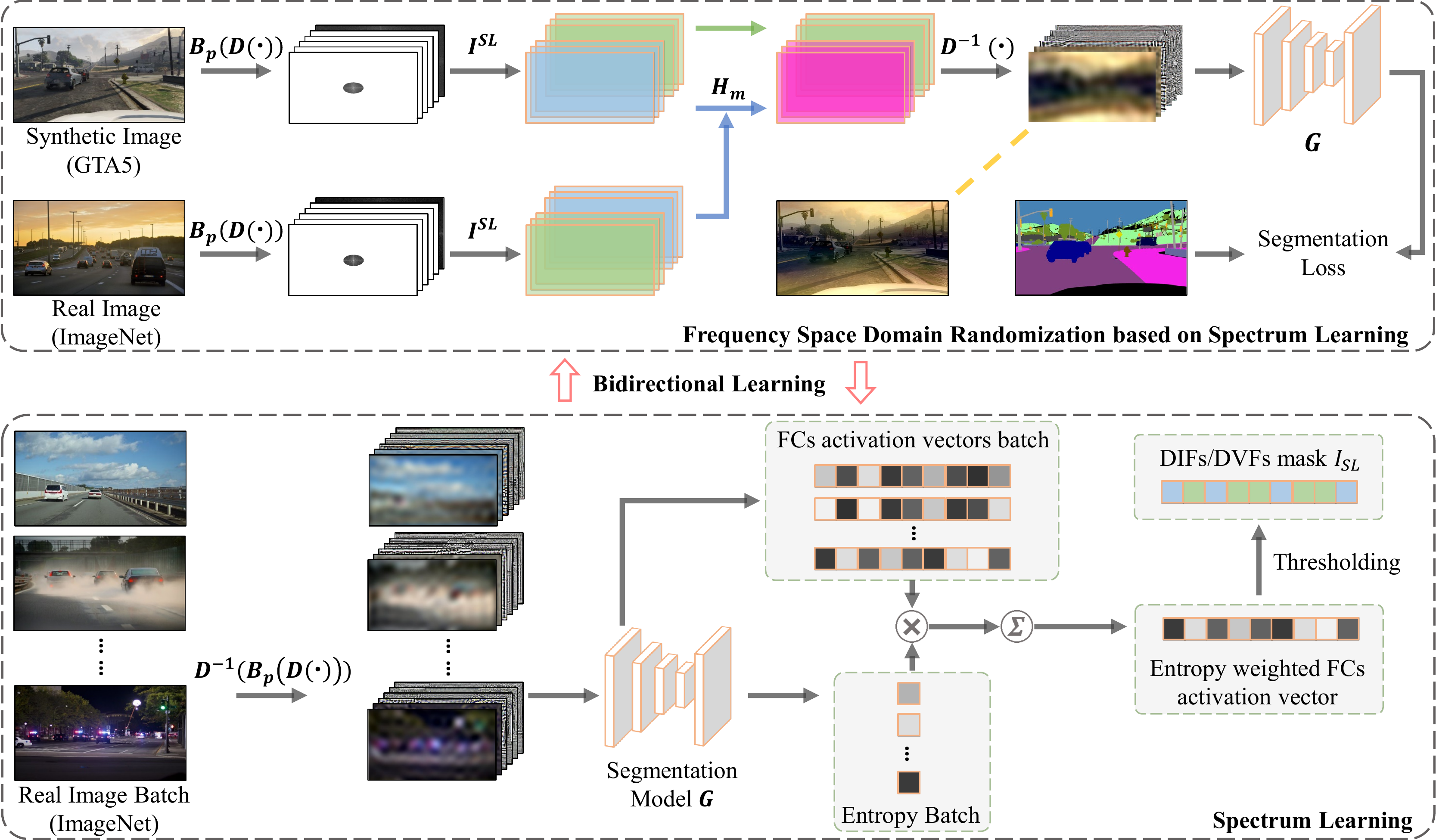}}
\vspace{-2pt}
\caption{
Overview of spectrum learning based frequency space domain randomization (FSDR-SL): FSDR-SL is a bidirectional learning framework that consists of two alternative learning processes, namely, Spectrum Learning (bottom part) and Frequency Space Domain Randomization (top part). In Spectrum Learning, we first decompose a batch of real images into spatial representation of FCs via $\mathcal{D}^{-1}(\mathcal{B}_{p}(\mathcal{D}(\cdot)))$ and then feed them to the segmentation model $G$ to identify DIFs and DVFs based on FCs activation vectors and corresponding prediction entropy. The learnt information is recorded in a binary mask vector $I^{SL}$. In FSDR, we first transform synthetic and real images into frequency space via $\mathcal{B}_{p}(\mathcal{D}(\cdot))$ and use the learnt $I^{SL}$ to randomize DVFs of the synthetic image conditioned on the reference real image via histogram matching (\ie, $\mathcal{H}_{m}$). We then employs $\mathcal{D}^{-1}(\cdot)$ to transform the randomized synthetic images into spatial representation and use them to optimize the segmentation model via segmentation loss. Note the reference real image in FSDR is sampled from the \textit{Real Image Batch} in Spectrum Learning, which forms a bidirectional learning framework. $\mathcal{D}(\cdot)$ and $\mathcal{D}^{-1}(\cdot)$ denotes discrete cosine transform (DCT) and inverse DCT; $\mathcal{B}_{p}(\cdot)$ represents a band-pass filter that decomposes the input into multiple FCs; Green boxes denote DIFs and green arrow represents keeping DIFs unchanged; Blue boxes denote DVFs and blue arrows represent randomizing DVFs; Gray arrows denote regular data flow; Yellow dash line indicates full-spectrum visualization of the randomized image.
}
\label{fig:stru}
\end{figure*}

\subsection{Spectrum Analysis in Domain Randomization}
This subsection describes the spectrum analysis based frequency space domain randomization (FSDR-SA) by using the ``$I^{SA}$". Different from existing domain randomization \cite{yue2019domain} that employs GANs for image stylization, we adopt histogram matching \cite{pizer1987adaptive} for online image translation. Specifically, we adjust the frequency space coefficient histogram of source images to be similar to that of the reference images by matching their cumulative density function. With ``$I^{SA}$", this histogram matching based randomization applies to DVFs of source images only without affecting DIFs. It adds little extra parameters and computation, and is much more efficient than GAN based translation. The top part of Figure \ref{fig:stru} illustrates how FSDR-SA works by simply replacing $I^{SL}$ with $I^{SA}$.

Given a source image $x_{s} \subset X_{s}$, the corresponding pixel-level label $y_{s} \subset Y_{s}$, the domain invariant FCs mask vector $I^{SA}$ and a ImageNet image $x_{img} \subset X_{\text{ImageNet}}$ as randomization reference, we first transform $x_s$ and $x_{img}$ into frequency space and decompose them by $f_{s} = \mathcal{B}_{p}(\mathcal{D}(x_{s})), f_{img} =\mathcal{B}_{p}(\mathcal{D}(x_{img}))$. We then randomize DVFs of $f_{s}$ 
with that of $f_{img}$ by adjusting the histogram of $f_{s}(1-I^{SA})$ to match that of $f_{img}(1-I^{SA})$. The FSDR-SA function $\mathcal{R}^{SA}(x_{s};x_{ref})$ can be defined by:
\begin{equation}
\mathcal{R}^{SA}(x_{s};x_{img}) = \\ \mathcal{D}^{-1}(\mathcal{H}_m(f_{s}(1-I^{SA}), f_{img}(1-I^{SA}))),
\end{equation}
where $\mathcal{H}_m()$ is the histogram matching function that adjusts the histogram of first input to match that of second.

The training loss of the FSDR-SA can be defined by:
\begin{equation}
\mathcal{L}_{SA} = l(G(\mathcal{R}^{SA}(x_{s};x_{img})), y_{s})
\end{equation}

\subsection{FSDR based on Spectrum Learning}
This section describes our spectrum learning based frequency space domain randomization (FSDR-SL) that identifies DIFs and DVFs via iterative learning, which enables dynamical and adaptive FSDR. We implement FSDR-SL with entropy~\cite{shannon1948mathematical} that was widely adopted in different tasks such as semi-supervised learning \cite{zhu2005semi,grandvalet2005semi,springenberg2015unsupervised}, clustering~\cite{jain2017subic,jain2018learning}, domain adaptation \cite{vu2019advent,Zou_2019_ICCV}, etc.

`Entropy' works by measuring class overlap~\cite{grandvalet2005semi,zhu2005semi,vu2019advent,Zou_2019_ICCV}, \ie, the prediction entropy decreases as classes overlap increases~\cite{castelli1996relative,o1978normal}. Leveraging this property, FSDR-SL identifies DIFs and DVFs according to the prediction entropy of reference images. Specifically, FSDR-SL is trained by using the decomposed multi-channel FCs of source images. If the trained model produces low entropy ($i.e.$, high confidence) predictions for a real target image, it indicates that the activated FCs of the target image are predictive with good semantic invariance across domains. In such cases, the employed FCs are identified as DIFs and randomization will be applied to other FCs to encourage the network to generate and learn invariant features during the iterative training process. Otherwise, no actions are taken as it is not clear whether non-activated FCs are semantically variant/uncorrelated or invariant/correlated. Note the image translation process is the same as that in FSDR-SA.

The idea of FSDR-SL is quite similar to that of self-training that either takes low-entropy/high-confidence predictions \cite{zou2018unsupervised,Zou_2019_ICCV}) as pseudo labels or directly minimizes the entropy of high-entropy/low-confidence predictions \cite{vu2019advent}. FSDR-SL reduces the overall prediction entropy by preserving low-entropy FCs while randomizing the rest FCs in domain randomization. Specifically, we first transform a batch of ImageNet images $x_{img}^{b} \in X_{\text{ImageNet}}$ into spatial representation of FCs $\mathcal{D}^{-1}(\mathcal{B}_{p}(\mathcal{D}(x_{img}^{b})))$ as shown in the bottom part of Fig.~\ref{fig:stru}. We then feed them to the segmentation model $G$ to identify DIFs and DVFs based on FCs activation vectors and corresponding prediction entropy. The learnt information is recorded in a binary mask vector $I^{SL}$ as follows:
\begin{align}
& E^{b}, \ A^{b} \leftarrow G(\mathcal{D}^{-1}(\mathcal{B}_{p}(\mathcal{D}(x_{img}^{b})))),\nonumber \\
& I^{SL} = RS(\sum_{b=1}^{B}(-A^{b}E^{b});p),\nonumber
\end{align}
where $E^{b}$ denotes a batch of averaged prediction entropy, $A^{b}$ denotes a batch of averaged input FC activation vectors, $B$ denotes the batch size, $RS$ denotes rank and select FCs with top $p$ portion entropy-weighted activation values as domain invariant FCs; $I^{SL}$ is a 192 length binary vector whose values record whether each FC is domain invariant (\ie, $I^{SL} = 1$) or variant (\ie, $I^{SL} = 0$). The \textit{Spectrum Learning} algorithm is included in appendix \textcolor{red}{A.2.}

Given a source image $x_{s} \subset X_{s}$, the corresponding pixel-level label $y_{s} \subset Y_{s}$ and a randomly picked ImageNet image $x_{img} \subset X_{\text{ImageNet}}$ as randomization reference, we first transform $x_s$ and $x_{img}$ to frequency space and decompose them to $f_{s} = \mathcal{B}_{p}(\mathcal{D}(x_{s}))$ and $f_{img} =\mathcal{B}_{p}(\mathcal{D}(x_{img}))$. The DVFs of $f_{s}$ (\ie, $\{f_{s}(i) | I^{SL}(i) = 0\}$) can then be randomized with that of $f_{img}$ via histogram matching. The FSDR-SL function $\mathcal{R}^{SL}_{inter}(x_{s};x_{img})$ can be defined by:
\begin{equation}
\mathcal{R}^{SL}(x_{s};x_{img}) = \mathcal{D}^{-1}(\mathcal{H}_m(f_{s}(1-I^{SL}), f_{img}(1-I^{SL}))),
\end{equation}
where $\mathcal{H}_m()$ denotes the histogram matching function as used in Eq. 4. The FSDR-RL training loss can thus be defined as follows:
\begin{equation}
\mathcal{L}_{SL} = l(G(\mathcal{R}^{SL}(x_{s};x_{img})), y_{s})
\end{equation}

Note the reference real image in FSDR is sampled from the Real Image Batch that have been performed Spectrum Learning. Once it is run out, the training goes back to Spectrum Learning process. FSDR-SL thus forms a bidirectional learning framework as shown in Figure \ref{fig:stru}, which consists of two alternative steps: 1) Perform spectrum learning with Real Image Batch on current model; 2) Perform FSDR with spectrum learnt reference images and update model. The FSDR-SL algorithm is available in appendix \textcolor{red}{A.3.}

While combining the spectrum analysis and spectrum learning, the overall training objective of the proposed FSDR can be defined as follows:
\begin{equation}
\min_{\theta} (\mathcal{L}_{orig} + \mathcal{L}_{SA} + \mathcal{L}_{SL})
\end{equation}

\section{Experiments}
This section presents evaluations of our proposed FSDR including datasets and implementation details, comparisons with the state-of-the-art, ablation studies, and discussion, more details to be described in the ensuing subsections.

\subsection{Datasets}
We evaluate FSDR over two challenging unsupervised domain generalization tasks GTA5$\rightarrow$\{Cityscapes, BDD, Mapillary\} and SYNTHIA$\rightarrow$\{Cityscapes, BDD, Mapillary\} that involve two synthetic source datasets and three real target datasets. GTA5 consists of $24,966$ high-resolution synthetic images, which shares 19 classes with Cityscapes, BDD, and Mapillary. SYNTHIA consists of $9,400$ synthetic images, which shares 16 classes with the three target datasets. Cityscapes, BDD, and Mapillary consist of $2975$, $7000$, and $18000$ real-world training images and $500$, $1000$, and $2000$ validation images.
Note we did not use any target data in training, but just a small subset of ImageNet~\cite{deng2009imagenet} as references for “stylizing/randomizing” the source-domain images as in \cite{yue2019domain}.
The evaluations were performed over the official validation data of the three target-domain datasets as in \cite{yue2019domain,pan2018two,zou2018unsupervised,tsai2018learning,vu2019advent,Pan_2020_CVPR}.

\subsection{Implementation Details}
We use ResNet101 \cite{he2016deep} or VGG16 \cite{simonyan2014very} (pre-trained using ImageNet \cite{deng2009imagenet}) with FCN-8s \cite{long2015fully} as the segmentation model $G$. The optimizer is Adam with momentum of $0.9$ and $0.99$. The learning rate is $1e-5$ initially and decreased with `step' learning rate policy with step size of 5000 and $\gamma = 0.1$. 
For all backbones, we modify input channels from 3 (\ie, RGB channels) to 192 (\ie, the frequency-wise multi-channel spatial representation, where one full spectrum is decomposed into 64 channels) for ``Spectrum Learning" that identifies DVFs and DIFs adaptively. This modification adds a little extra parameters and computation for the whole networks - it only adds about $0.0786\%$ and $1.33\%$ extra parameter and computation for VGG-16 and ResNet-101, respectively.
During training, we train FSDR-SL and FSDR without using $\mathcal{L}_{SL}$ in the first epoch (to avoid very noisy predictions at the initial training stage) and then use all losses in the ensuing epochs.

\renewcommand\arraystretch{1.1}
\begin{table}[t]
\caption{
Ablation study for the domain generalization task GTA $\rightarrow$ \{Cityscapes, Mapillary and BDD\} (using ResNet-101 as backbone) in mIoU. Losses $L_{orig}$, $L_{SA}$, and $L_{SL}$ are defined in Eq. 1, 5, and 7, respectively.
}
\centering
\begin{small}
\begin{tabular}{ccccccc}
\hline
\hline
\multicolumn{1}{c|}{\multirow{2}{*}{\textbf{Method}}} &\multirow{2}{*}{$\mathcal{L}_{orig}$} &\multicolumn{1}{c}{\multirow{2}{*}{$\mathcal{L}_{SA}$}} &\multicolumn{1}{c|}{\multirow{2}{*}{$\mathcal{L}_{SL}$}}& \multicolumn{3}{c}{\textbf{mIoU}}
\\\cline{5-7} 
\multicolumn{1}{c|}{}  & &  &\multicolumn{1}{c|}{}  & \multicolumn{1}{c}{City.} & \multicolumn{1}{c}{Mapi.} & \multicolumn{1}{c}{BDD} 
\\\hline
Baseline &\multicolumn{1}{c}{\checkmark}& & &33.4 & 27.9  &27.3\\
FSDR-SA &\multicolumn{1}{c}{\checkmark} &\multicolumn{1}{c}{\checkmark} & &42.1 & 39.2  &37.8\\
FSDR-SL &\multicolumn{1}{c}{\checkmark} & &\multicolumn{1}{c}{\checkmark} &43.6 &42.1 &40.1\\
FSDR &\multicolumn{1}{c}{\checkmark} &\multicolumn{1}{c}{\checkmark} &\multicolumn{1}{c}{\checkmark} &44.8 & 43.4  &41.2\\
\hline
\end{tabular}
\end{small}
\label{tab:abla}
\end{table}

\subsection{Ablation Studies}
We examine different FSDR designs to find out how they contribute to the network generalization in semantic segmentation. As shown in Table \ref{tab:abla}, we trained four models over the UDG task GTA5$\rightarrow$\{Cityscapes, BDD, Mapillary\}: 1) \textit{Baseline} that is trained with $L_{orig}$ without randomization as shown in Section \textcolor{red}{3.1}, 2) \textit{FSDR-SA} that is trained with spectrum analysis based randomization using $L_{SA}$ and $L_{orig}$, 3) \textit{FSDR-SL} that is trained with spectrum learning based randomization using $L_{orig}$ and $L_{SL}$, and 4) \textit{FSDR} that is trained with $L_{orig}$, $L_{SA}$, and $L_{SL}$.

We applied the four models to the validation data of Cityscapes, BDD, and Mapillary and Table \ref{tab:abla} shows experimental results. It can be seen that \textit{Baseline} trained with the GTA data only (\ie, $L_{orig}$) does not performs well due to domain bias. \textit{FSDR-SA} and \textit{FSDR-SL} outperform \textit{Baseline} by large margins, demonstrating the importance of preserving domain invariant features in domain randomization for training domain generalizable models. Additionally, \textit{FSDR-SL} outperforms \textit{FSDR-SA} clearly which is largely due to the adaptive and iterative spectrum learning that enables the dynamic and adaptive FSDR and a bidirectional learning framework.
Further, \textit{FSDR} performs the best consistently. It shows that \textit{FSDR-SA} and \textit{FSDR-SL} are complementary where the static spectrum analysis forms certain bases for effective and stable spectrum leaning while training domain generalizable models.

\begin{table}[t]
\centering
\caption{Domain generalization performance for the task GTA $\rightarrow$ \{\textcolor{red}{C}ityscapes, \textcolor{red}{M}apillary, and \textcolor{red}{B}DDS\} in mIoU. ``w/ Tgt" labels the methods that train models with (\textcolor{red}{\cmark} \ie, domain adaptation) and without (\textcolor{green}{\xmark} \ie, domain generalization) accessing target-domain data in Cityscapes.}
\vspace{2mm}
\label{Tab: GTAseg}
\resizebox{\columnwidth}{!} {
\begin{tabular}{c|l||c|ccc|c}

\toprule
\hline
Net. & Method & \multicolumn{1}{c|}{\begin{tabular}[c]{@{}c@{}}w/\\ Tgt\end{tabular}} & C & M & B & Mean \\ \hline

\multicolumn{1}{c|}{\multirow{8}{*}{{\begin{tabular}[c]{@{}c@{}}Res\\ Net\\ 101\end{tabular}}}}
& CBST~\cite{zou2018unsupervised}  &\textcolor{red}{\cmark} &44.9 &40.3 &40.5 &41.9 \\
\multicolumn{1}{l|}{}                        & AdaSeg.~\cite{tsai2018learning}       &\textcolor{red}{\cmark}     &41.4     & 38.3 &36.2 &38.6     \\
\multicolumn{1}{l|}{}                        & MinEnt~\cite{vu2019advent}       &\textcolor{red}{\cmark}     &42.3     & 38.5 &34.4 &38.4     \\
\multicolumn{1}{l|}{}                        & FDA~\cite{yang2020fda}       &\textcolor{red}{\cmark}     &45.0     & 39.6 &38.1 &40.9    \\
\multicolumn{1}{l|}{}                        & IDA~\cite{Pan_2020_CVPR}       &\textcolor{red}{\cmark}     &\textbf{45.1}     &39.4 &37.6 &40.7      \\\cline{2-7}
&IBN-Net~\cite{pan2018two} &\textcolor{green}{\xmark} &40.3 &35.9 &35.6 &37.2 \\
&DRPC~\cite{yue2019domain} &\textcolor{green}{\xmark} &42.5 &38.0 &38.7 &39.8 \\
& \textbf{Ours (FSDR)} &\textcolor{green}{\xmark} &44.8 &\textbf{43.4} & \textbf{41.2} &\textbf{43.1} \\\hline
\multirow{8}{*}{{\begin{tabular}[c]{@{}c@{}}VGG\\ 16\end{tabular}}}                     
  \\
                                             & CBST~\cite{zou2018unsupervised} &\textcolor{red}{\cmark}  &38.1 & 34.6  &33.9 &35.5  \\
                                             & AdaSeg.~\cite{tsai2018learning}  &\textcolor{red}{\cmark}  &35.0  & 32.6 &31.3 &33.0  \\
                     & MinEnt~\cite{vu2019advent}         &\textcolor{red}{\cmark}     &32.8      &30.7  &29.5 &31.0 \\
                     & FDA~\cite{yang2020fda}         &\textcolor{red}{\cmark}     &37.9      &33.8  &32.1 &34.6 \\
                                             & IDA~\cite{Pan_2020_CVPR}         &\textcolor{red}{\cmark}     &\textbf{38.5}      & 34.2 &32.3 &35.0 \\  \cline{2-7}
                                                        
&IBN-Net~\cite{pan2018two} &\textcolor{green}{\xmark} &34.8 &31.0 &30.4 &32.0 \\
& DRPC~\cite{yue2019domain} &\textcolor{green}{\xmark} &36.1 &32.3 &31.6 &33.3 \\
& \textbf{Ours (FSDR)} &\textcolor{green}{\xmark} &38.3 &\textbf{37.6} &\textbf{34.4} &\textbf{37.1} \\ \hline
\end{tabular}}
\end{table}

\subsection{Comparisons with the State-of-Art}

\begin{figure*}[t]
\begin{tabular}{p{3cm}<{\centering} p{3cm}<{\centering} p{3cm}<{\centering} p{3cm}<{\centering} p{3cm}<{\centering}}
\raisebox{-0.5\height}{\includegraphics[width=1.1\linewidth,height=0.55\linewidth]{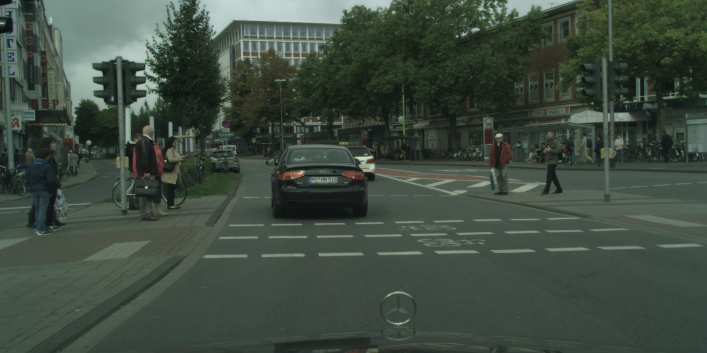}}
 & \raisebox{-0.5\height}{\includegraphics[width=1.1\linewidth,height=0.55\linewidth]{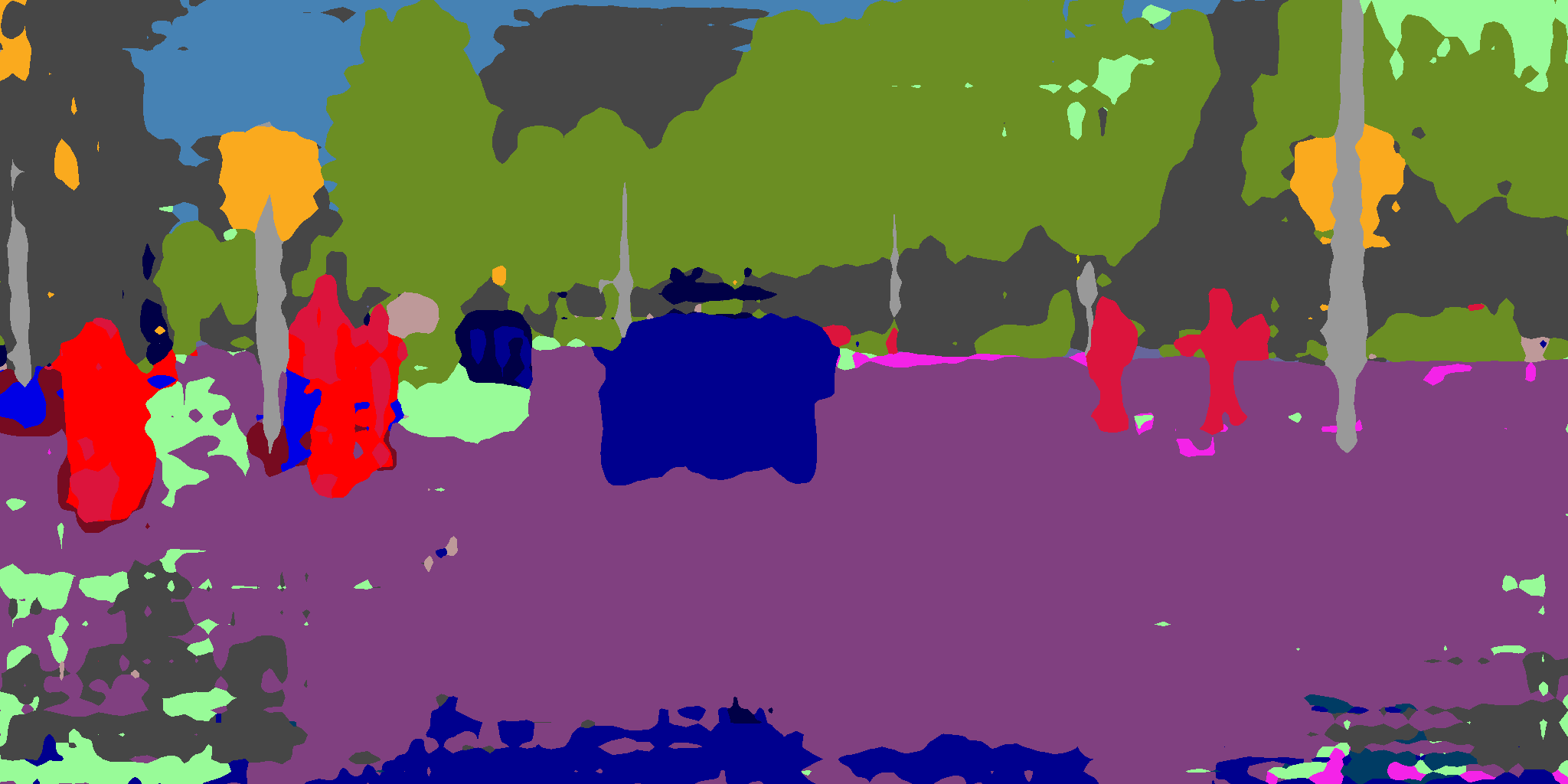}}
& \raisebox{-0.5\height}{\includegraphics[width=1.1\linewidth,height=0.55\linewidth]{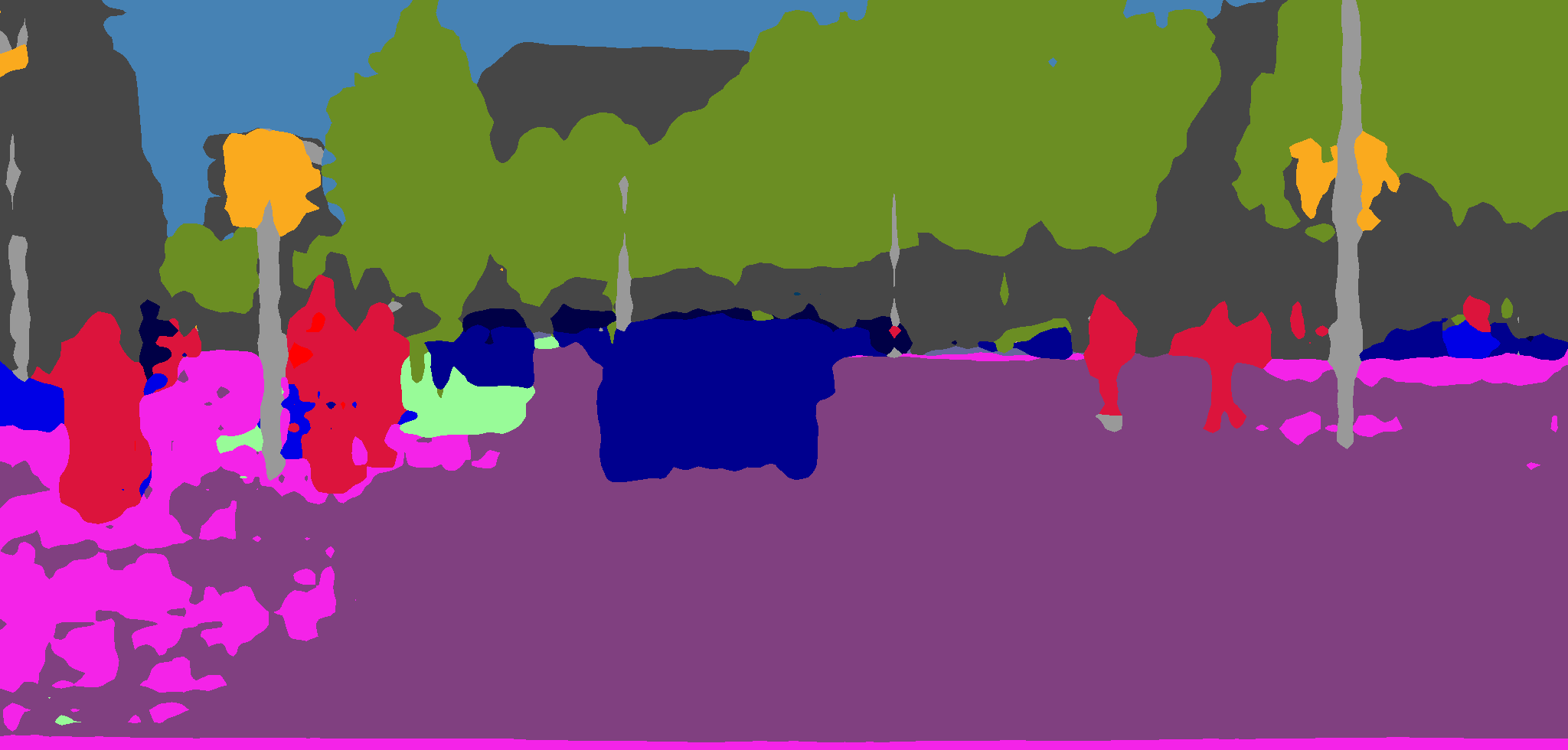}}
& \raisebox{-0.5\height}{\includegraphics[width=1.1\linewidth,height=0.55\linewidth]{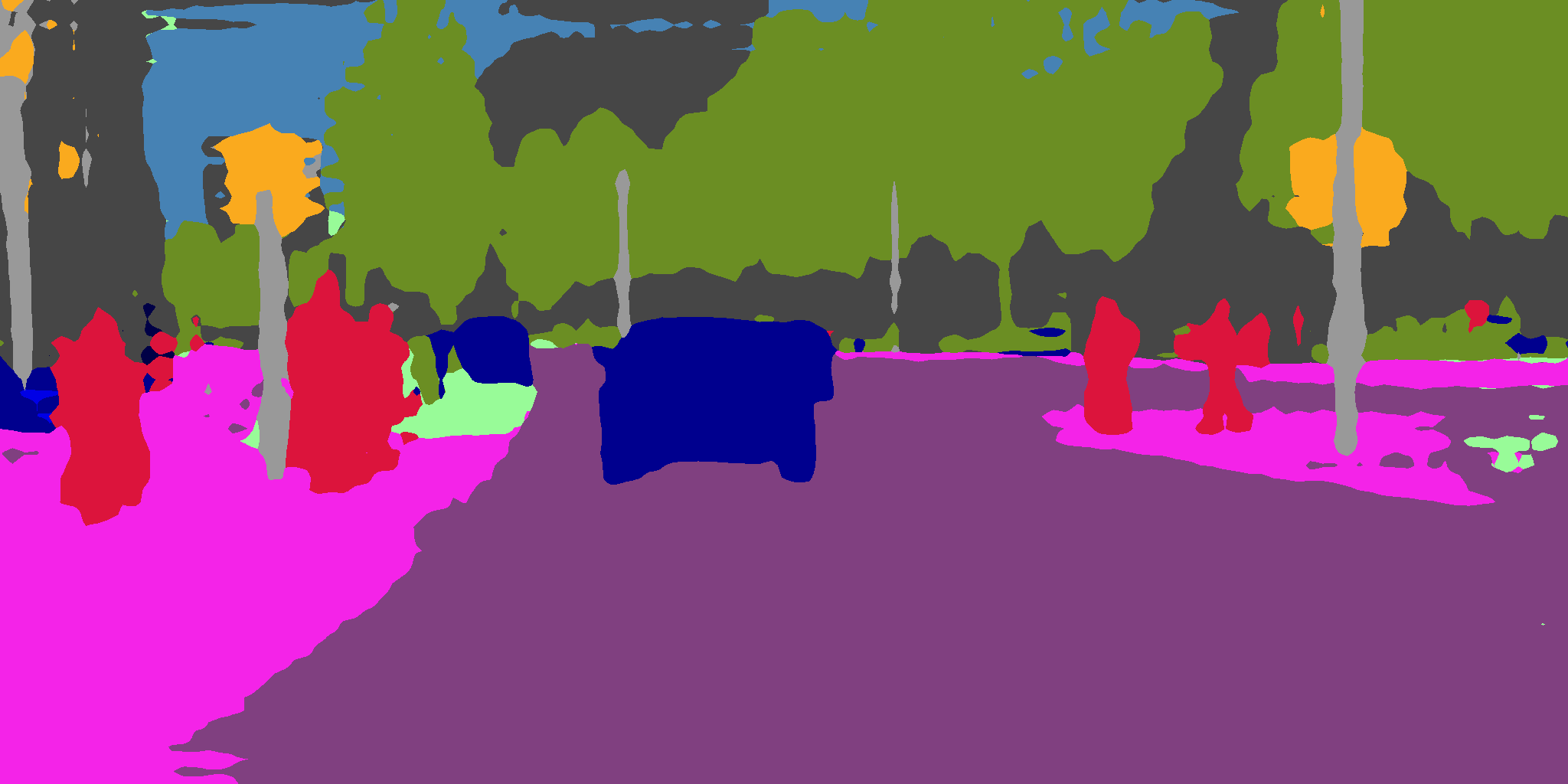}}
& \raisebox{-0.5\height}{\includegraphics[width=1.1\linewidth,height=0.55\linewidth]{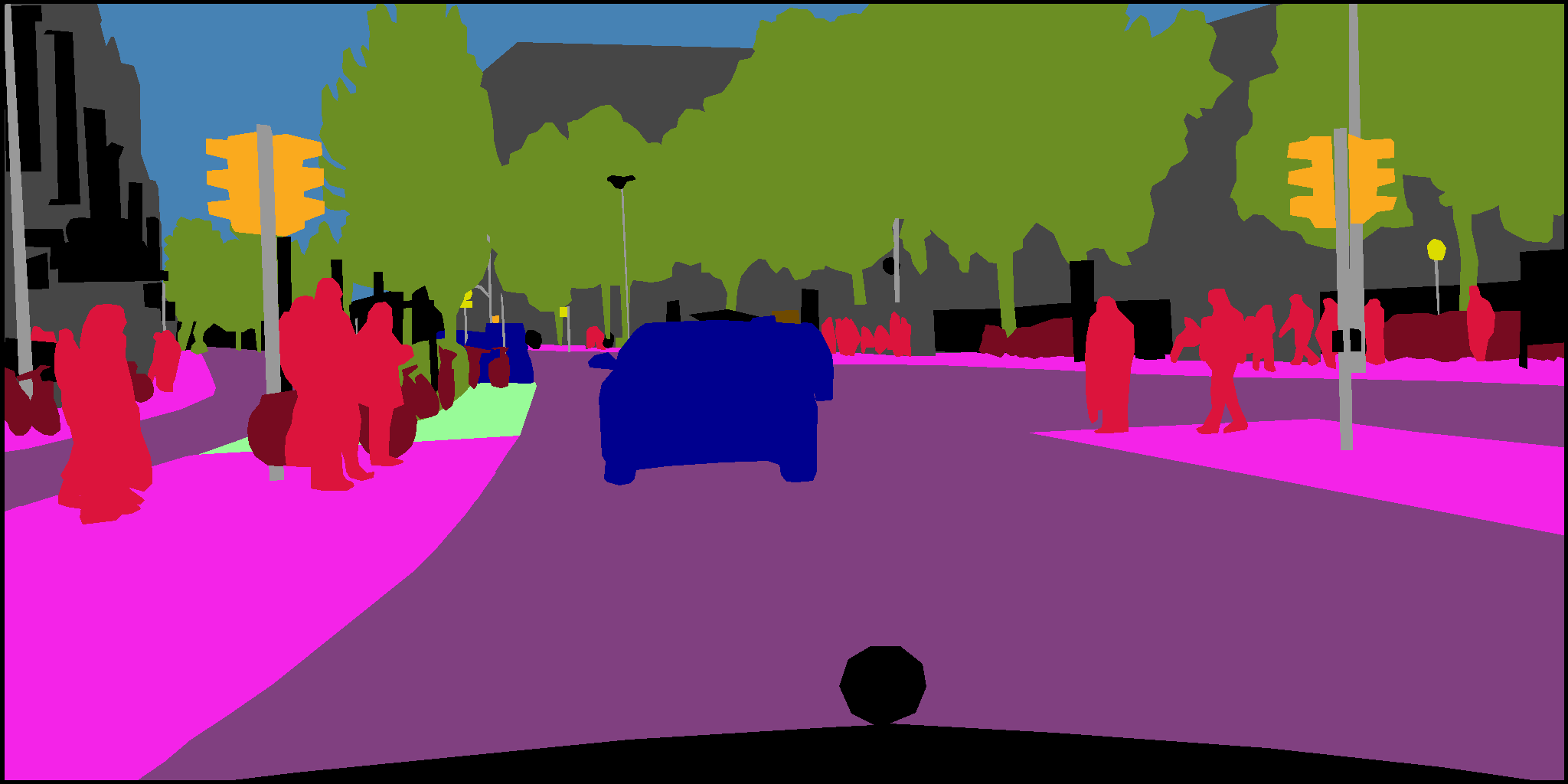}}
\vspace{-2.5 pt}
\\
\raisebox{-0.5\height}{\includegraphics[width=1.1\linewidth,height=0.55\linewidth]{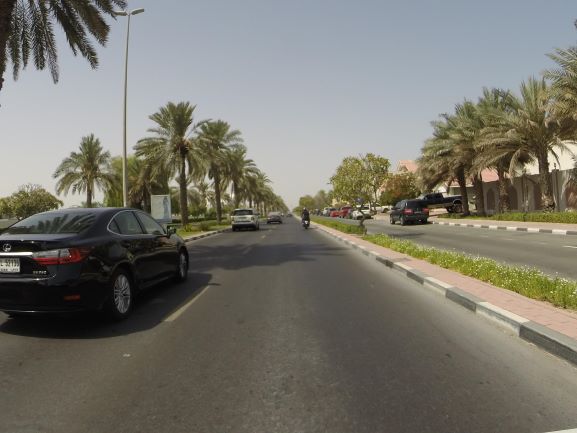}}
 & \raisebox{-0.5\height}{\includegraphics[width=1.1\linewidth,height=0.55\linewidth]{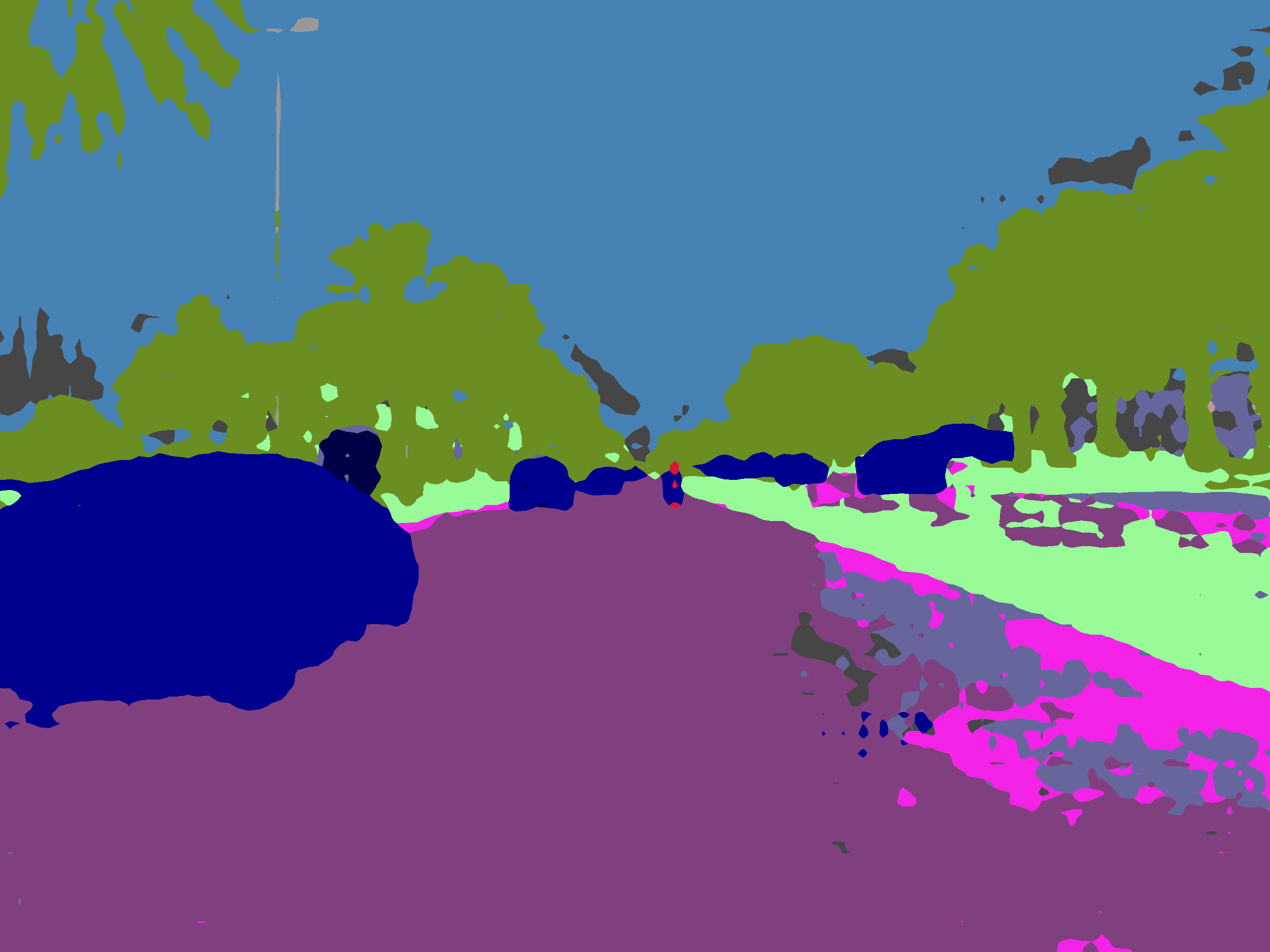}}
& \raisebox{-0.5\height}{\includegraphics[width=1.1\linewidth,height=0.55\linewidth]{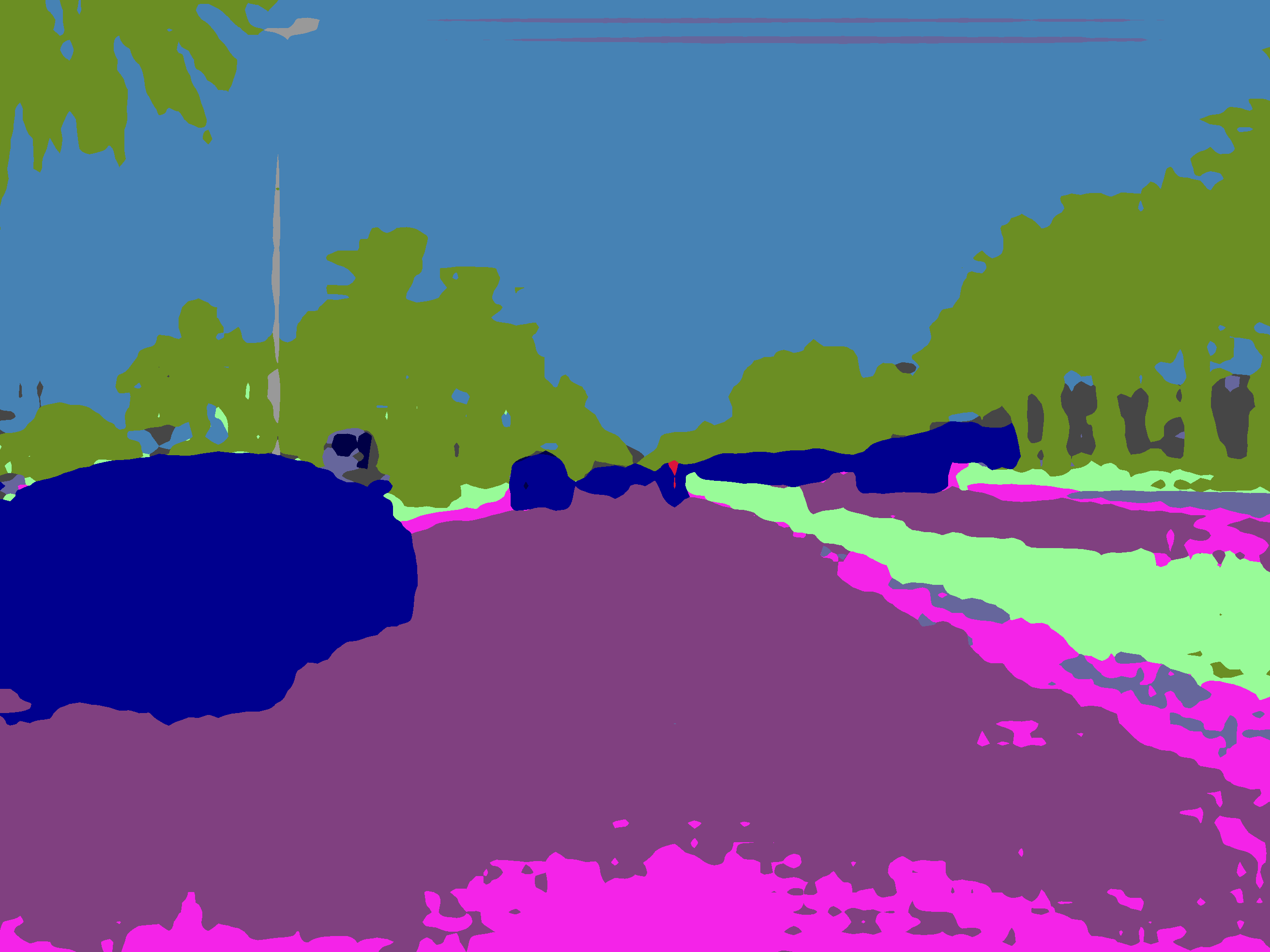}}
& \raisebox{-0.5\height}{\includegraphics[width=1.1\linewidth,height=0.55\linewidth]{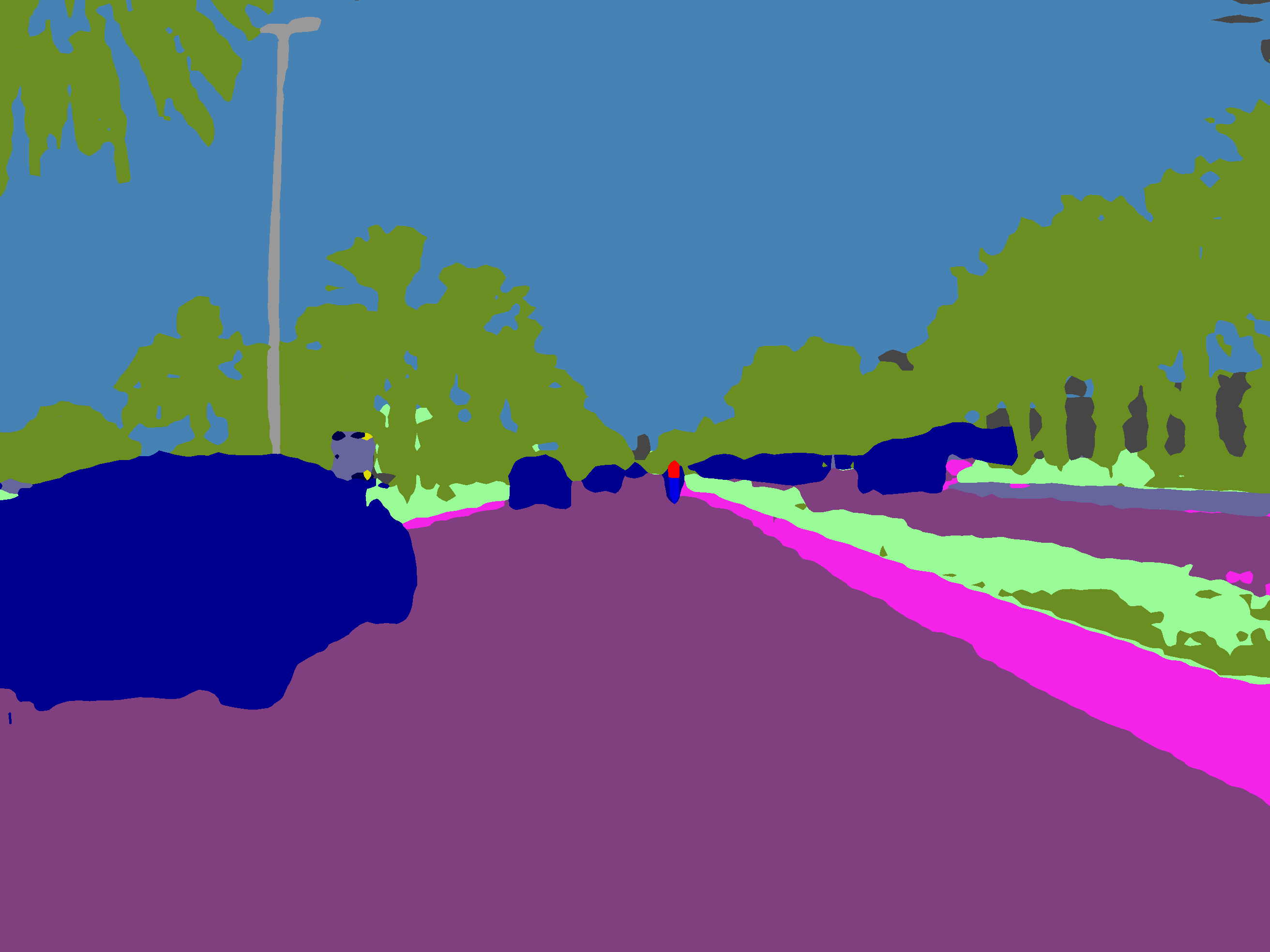}}
& \raisebox{-0.5\height}{\includegraphics[width=1.1\linewidth,height=0.55\linewidth]{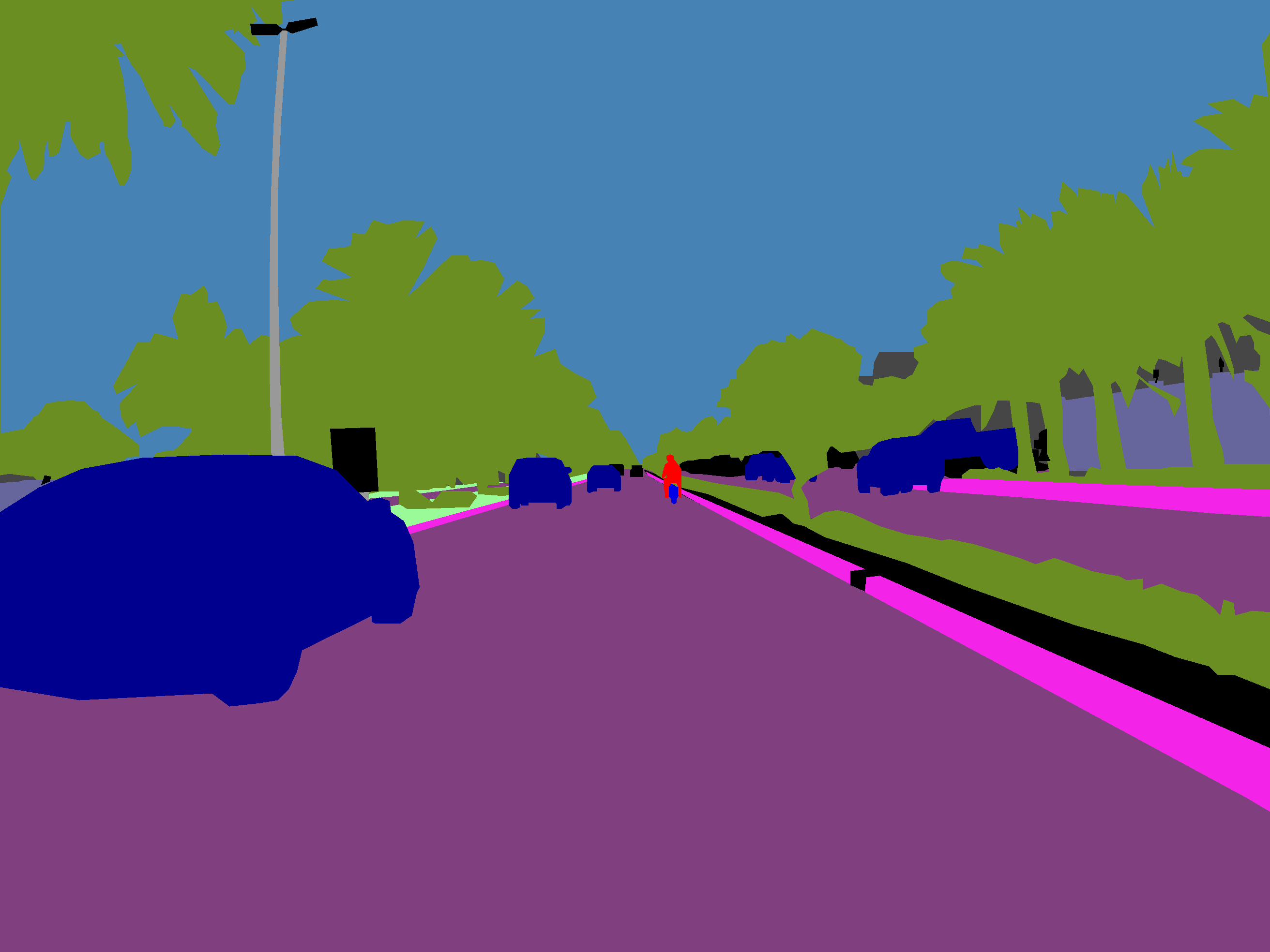}}
\vspace{-2.5 pt}
\\
\raisebox{-0.5\height}{\includegraphics[width=1.1\linewidth,height=0.55\linewidth]{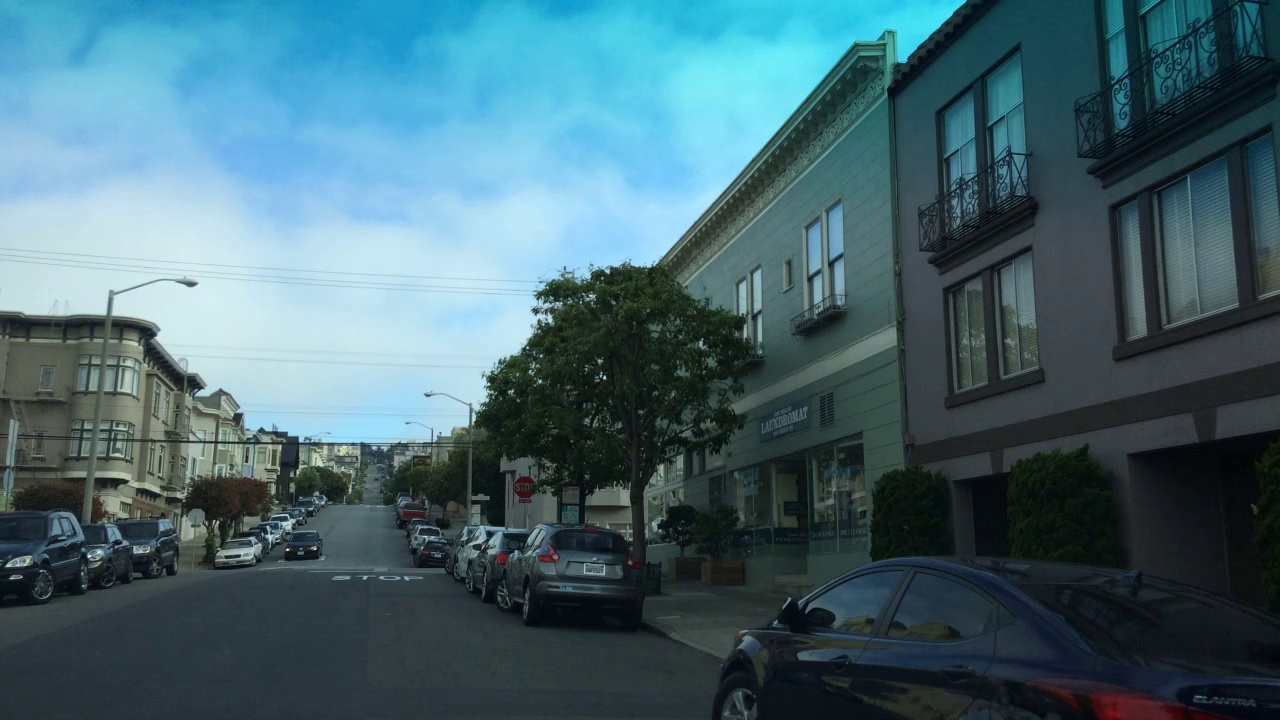}}
 & \raisebox{-0.5\height}{\includegraphics[width=1.1\linewidth,height=0.55\linewidth]{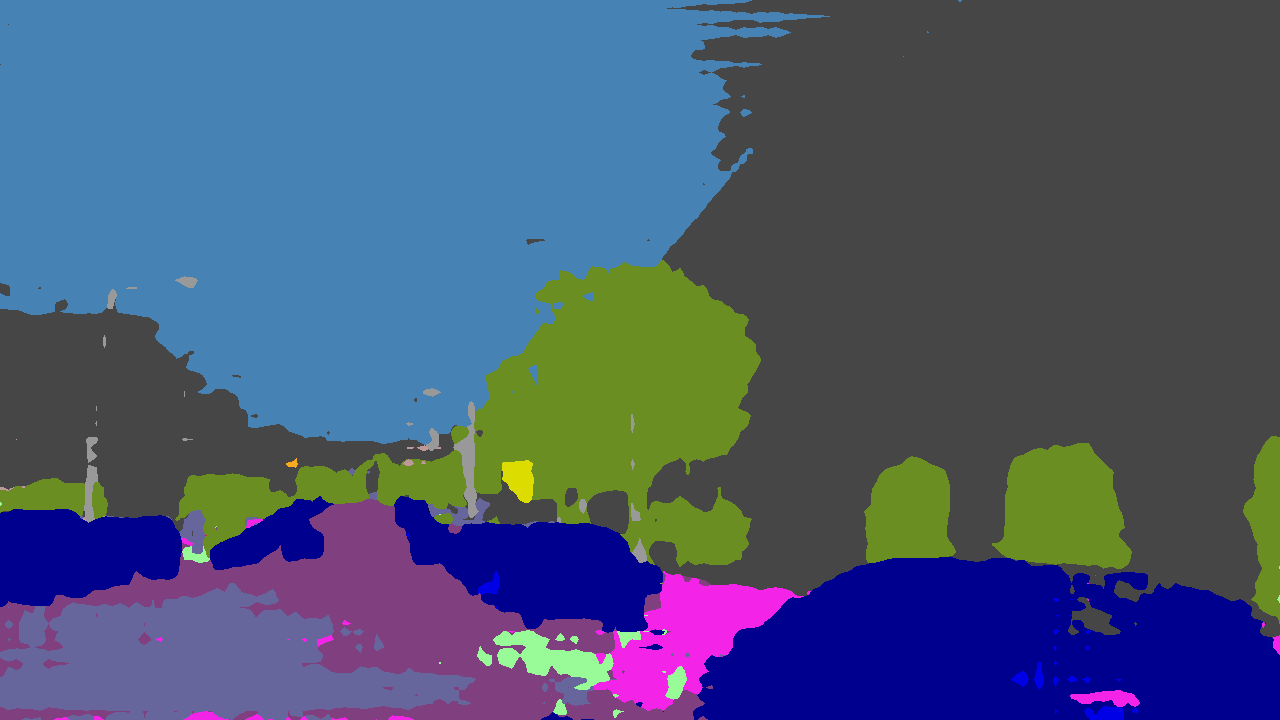}}
& \raisebox{-0.5\height}{\includegraphics[width=1.1\linewidth,height=0.55\linewidth]{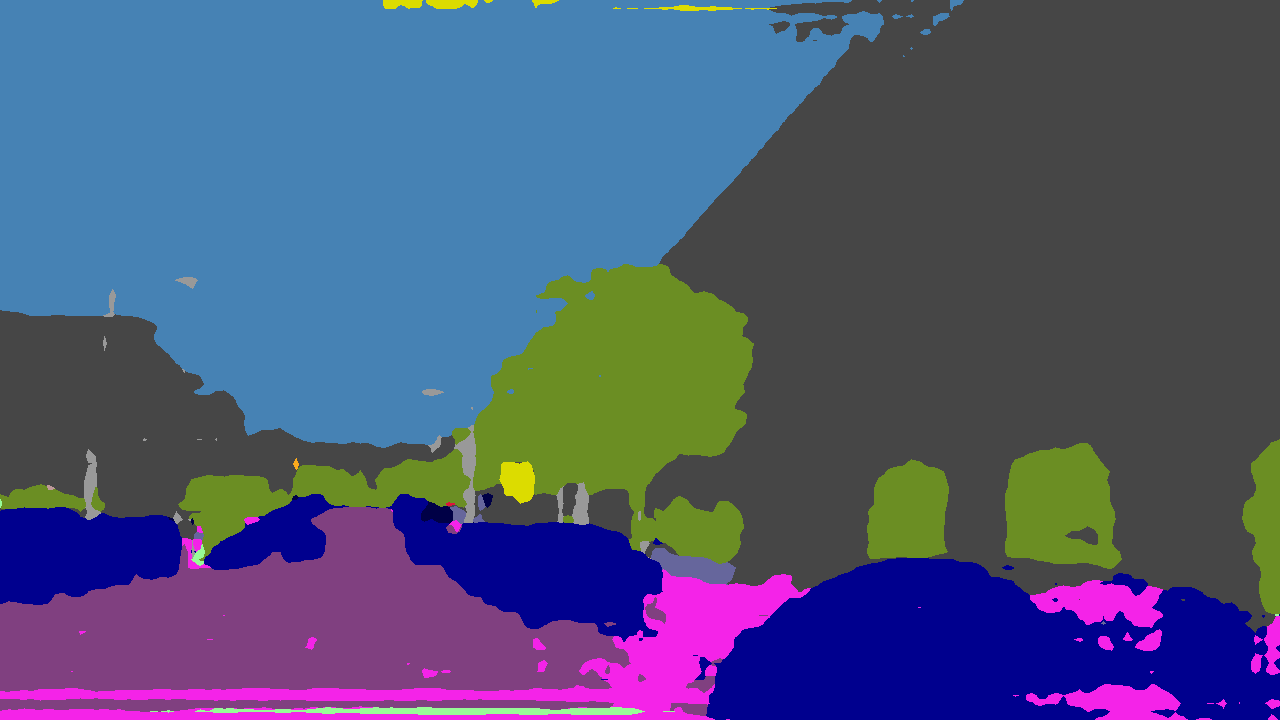}}
& \raisebox{-0.5\height}{\includegraphics[width=1.1\linewidth,height=0.55\linewidth]{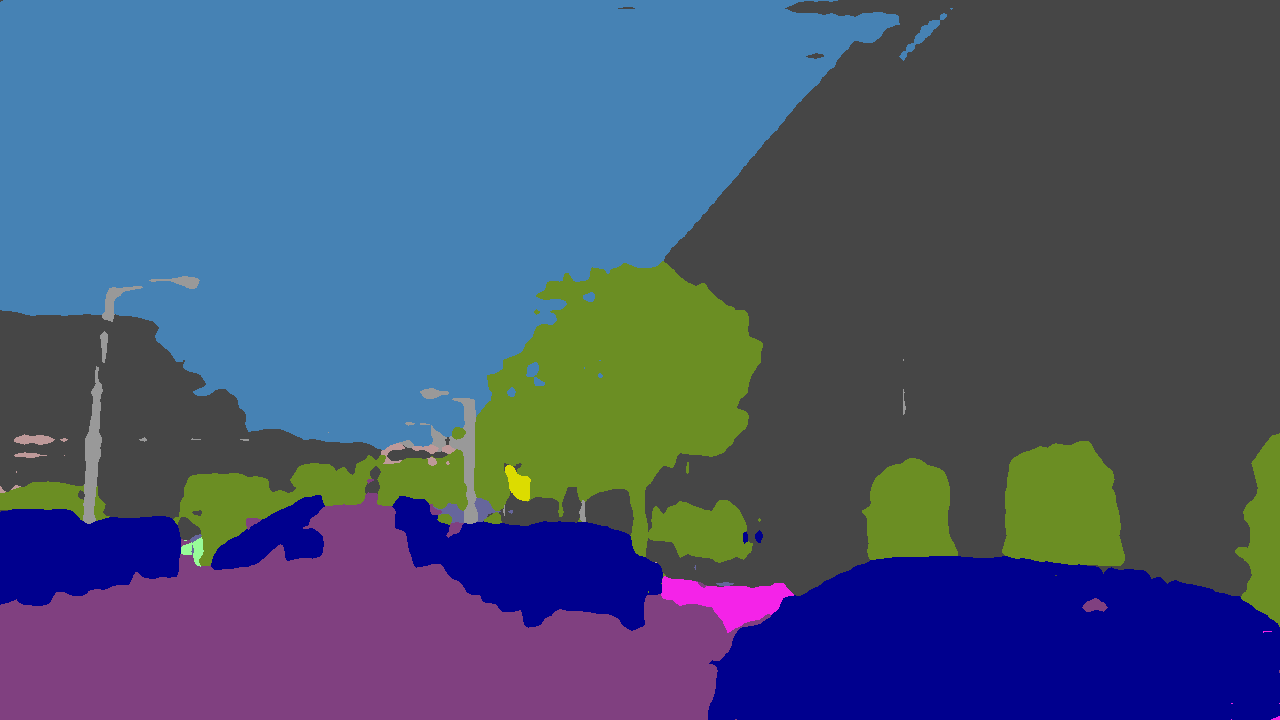}}
& \raisebox{-0.5\height}{\includegraphics[width=1.1\linewidth,height=0.55\linewidth]{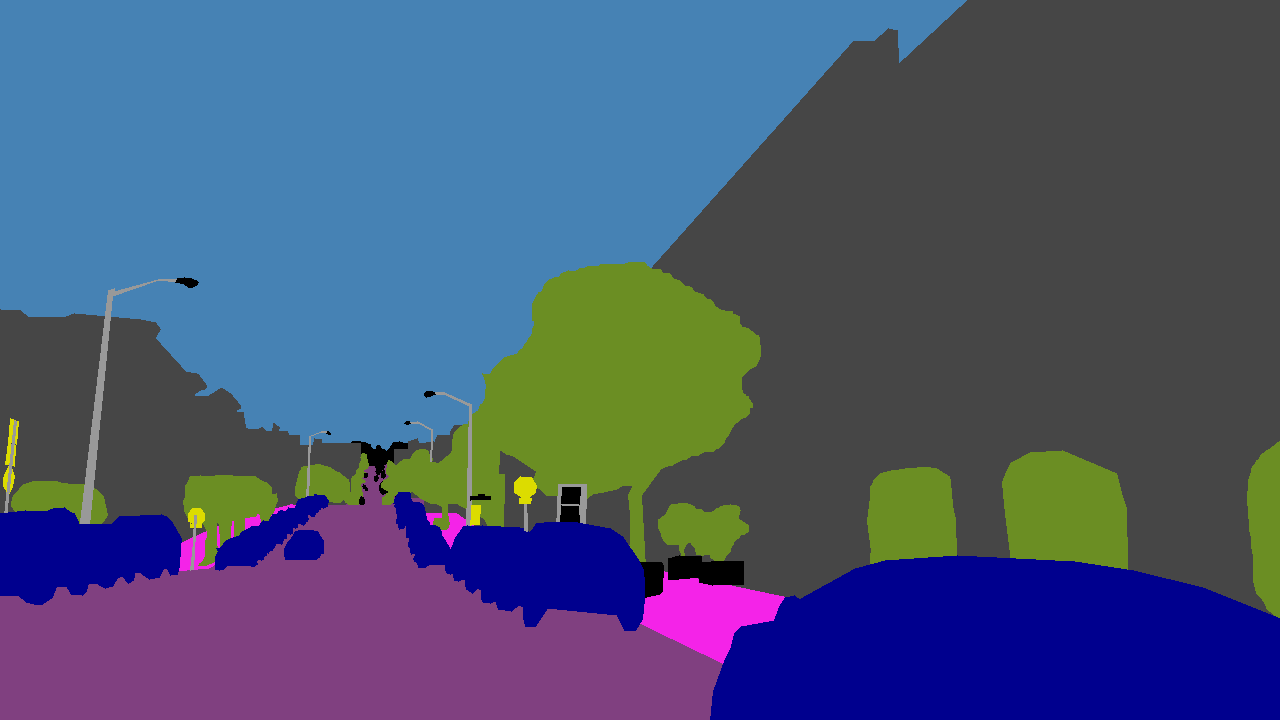}}
\vspace{-5.5 pt}
\\
\raisebox{-0.5\height}{Target Image}
 & \raisebox{-0.5\height}{Baseline}
& \raisebox{-0.5\height}{DRPC}
& \raisebox{-0.5\height}{\textbf{Ours (FSDR)}}
& \raisebox{-0.5\height}{Ground Truth}
\\
\end{tabular}
\vspace{2.5 pt}
\caption{
Qualitative illustration of domain generalizable semantic segmentation for GTA5 to Cityscapes (1st row), Mapillary (2nd row), and BDD (3rd row). FSDR preserves domain invariant information during domain randomization which produces better semantic segmentation especially around edge/class-transition area. As a comparison, DRPC~\cite{yue2019domain} does not isolate and preserve domain invariant feature which leads to sub-optimal segmentation.}
\label{fig:results}
\end{figure*}

We compared FSDR with a number of state-of-the-art UDG methods as shown in Tables \ref{Tab: GTAseg} and \ref{Tab: SYNTHIAseg} (highlighted by \textcolor{green}{\xmark}). The comparisons were performed over two tasks GTA5$\rightarrow$\{Cityscapes, BDD, Mapillary\} and SYNTHIA$\rightarrow$\{Cityscapes, BDD, Mapillary\} where two network backbones ResNet101 and VGG16 were evaluated for each task. As the two tables show, FSDR outperforms all state-of-the-art UDG methods clearly and consistently across both tasks and two different network backbones. The superior segmentation performance is largely attributed to the proposed frequency space domain randomization that identifies and keeps DIFs unchanged and randomizes DVFs only. Without identifying and preserving DIFs, state-of-the-art methods \cite{yue2019domain,goodfellow2014generative} tend to over-randomize images which may degrade image structures and semantics and lead to sup-optimal network generalization. We provide the qualitative comparison in Figure \ref{fig:results}.

We also compared FSDR with several state-of-the-art UDA methods as shown in Tables \ref{Tab: GTAseg} and \ref{Tab: SYNTHIAseg} (highlighted by \textcolor{red}{\cmark}), where each method adapts to Cityscapes only and the adapted model is then evaluated over Cityscapes, Mapillary, and BDD. It can be seen that FSDR is even on par with the state-of-the-art UDA method which accesses target-domain data (i.e. Cityscapes) in training. In addition, FSDR performs more stably across the three target domains, whereas UDA methods \cite{zou2018unsupervised,vu2019advent,tsai2018learning,Pan_2020_CVPR} demonstrate much larger mIoU drops for unseen (in training) Mapillary and BDD data. We conjecture that the UDA models may over-fit to the seen Cityscapes data which leads to poor generalization for unseen Mapillary and BDD data.

\begin{table}[t]
\centering
\caption{Domain generalization performance for the task SYNTHIA $\rightarrow$ \{\textcolor{red}{C}ityscapes, \textcolor{red}{M}apillary, and \textcolor{red}{B}DDS\} in mIoU. ``w/ Tgt" labels the methods that train models with (\textcolor{red}{\cmark} \ie, domain adaptation) and without (\textcolor{green}{\xmark} \ie, domain generalization) accessing target-domain data in Cityscapes.}
\vspace{2mm}
\label{Tab: SYNTHIAseg}
\resizebox{\columnwidth}{!} {
\begin{tabular}{c|l||c|ccc|c}
\toprule
\hline
Net. & Method & \multicolumn{1}{c|}{\begin{tabular}[c]{@{}c@{}}w/\\ Tgt\end{tabular}} & C & M & B & Mean \\ \hline

\multicolumn{1}{c|}{\multirow{8}{*}{{\begin{tabular}[c]{@{}c@{}}Res\\ Net\\ 101\end{tabular}}}} 
& CBST~\cite{zou2018unsupervised}  &\textcolor{red}{\cmark} &41.4 &37.1 &\textbf{37.6} &38.7 \\
\multicolumn{1}{l|}{}                        & AdaSeg.~\cite{tsai2018learning}       &\textcolor{red}{\cmark}     &38.2     &36.1   &35.3 &36.5      \\
\multicolumn{1}{l|}{}                        & MinEnt~\cite{vu2019advent}       &\textcolor{red}{\cmark}     &38.1     &35.8   &35.5 &36.4     \\
\multicolumn{1}{l|}{}                        & FDA~\cite{yang2020fda}       &\textcolor{red}{\cmark}     &41.2     &36.1   &36.4 &37.9     \\
\multicolumn{1}{l|}{}                        & IDA~\cite{Pan_2020_CVPR}       &\textcolor{red}{\cmark}     &\textbf{41.7}     &36.5   &37.0 &38.4      \\\cline{2-7}
&IBN-Net~\cite{pan2018two} &\textcolor{green}{\xmark} &37.5 &33.7 &33.0 &34.7 \\
&DRPC~\cite{yue2019domain} &\textcolor{green}{\xmark} &37.6  &34.1 &34.3 &35.4 \\
& \textbf{Ours (FSDR)} &\textcolor{green}{\xmark} &40.8 &\textbf{39.6} &37.4 & \textbf{39.3} \\\hline
\multirow{8}{*}{{\begin{tabular}[c]{@{}c@{}}VGG\\ 16\end{tabular}}}                     
& CBST~\cite{zou2018unsupervised}  &\textcolor{red}{\cmark} &38.2 &33.5  &32.2 &34.6 \\
\multicolumn{1}{l|}{}                        & AdaSeg.~\cite{tsai2018learning}       &\textcolor{red}{\cmark}     &32.6     &30.3   &29.4 &30.8      \\
\multicolumn{1}{l|}{}                        & MinEnt~\cite{vu2019advent}       &\textcolor{red}{\cmark}     &31.4     &29.8   &28.9 &30.0      \\
\multicolumn{1}{l|}{}                        & FDA~\cite{yang2020fda}       &\textcolor{red}{\cmark}     &37.9     &33.1   &31.8 &34.2      \\
\multicolumn{1}{l|}{}                        & IDA~\cite{Pan_2020_CVPR}       &\textcolor{red}{\cmark}     &\textbf{38.6}     &34.2  &32.7 &35.1      \\\cline{2-7}
&IBN-Net~\cite{pan2018two} &\textcolor{green}{\xmark} &33.9 &31.1 &30.4 &31.8 \\
                                             & DRPC~\cite{yue2019domain} &\textcolor{green}{\xmark} &35.5 &32.2 &29.5 &32.4 \\
                                             & \textbf{Ours (FSDR)} &\textcolor{green}{\xmark} &37.9 &\textbf{37.2} &\textbf{34.1} &\textbf{36.4} \\ \hline
\end{tabular}}
\end{table}

\subsection{Discussion}

FSDR is \textit{complementary} with most existing domain adaptation and domain generlization networks, which can be easily incorporated into them with consistent performance boost but little extra parameters and computation. We evaluated this feature by incorporating FSDR into a number of domain adaptation and domain generalization networks as shown in Table \ref{tab:complement_study}. During training, we reconstruct the FSDR randomized multi-FC representation back to the full-spectrum representation with 3 channels (i.e. Ours*) for compatibility with the compared domain adaptation and generalization methods that work with full-spectrum three-channel images. As Table \ref{tab:complement_study} shows, the incorporation of FSDR $\{\mathcal{L}_{SA}, \mathcal{L}_{SL}\}$ improves the semantic segmentation of state-of-the-art networks consistently. As the incorporation of FSDR just includes a few losses without changing network structures, the inference has little extra parameters and computation once the model is trained.

\begin{table}
\centering
\caption{
FSDR is complementary with existing domain adaptation and generalization methods. For the task GTA $\rightarrow$ \{Cityscapes, Mapillary, and BDD\}, including FSDR (Ours*) improves the domain adaptation and generalization performance consistently. ``w/ Tgt" labels methods that train models with (\textcolor{red}{\cmark} i.e. domain adaptation) or without (\textcolor{green}{\xmark} \ie, domain generalization) accessing target data in Cityscapes.}
\vspace{1mm}
\resizebox{\columnwidth}{!}{
\begin{tabular}{l|c|cc|cc|cc} \toprule
                             &\multicolumn{1}{c|}{\multirow{2}{*}{{\begin{tabular}[c]{@{}c@{}}w/\\ Tgt\end{tabular}}}}  & \multicolumn{2}{c}{Cityscapes}  & \multicolumn{2}{c}{Mapillary} & \multicolumn{2}{c}{BDD} \\
                                 \cmidrule{3-8}
                                 & & \multicolumn{1}{c}{Base} & \multicolumn{1}{c}{+Ours*} & \multicolumn{1}{c}{Base} & \multicolumn{1}{c}{+Ours*} & \multicolumn{1}{c}{Base} & \multicolumn{1}{c}{+Ours*} \\ \hline
Adapt-SegMap \cite{tsai2018learning}  &\textcolor{red}{\cmark} &41.4& \textbf{45.6}& 38.3& \textbf{43.9} &36.2 & \textbf{41.9} \\
MinEnt \cite{vu2019advent} &\textcolor{red}{\cmark} &42.3	&\textbf{45.7}	&38.5	&\textbf{43.7}	&34.4	&\textbf{41.7}\\
CBST~\cite{zou2018unsupervised} &\textcolor{red}{\cmark} &44.9	&\textbf{46.8}	&40.3	&\textbf{44.3}	&40.5	&\textbf{42.8} \\
FDA \cite{yang2020fda} &\textcolor{red}{\cmark} &45.0	&\textbf{46.1}	&39.6	&\textbf{44.2}	&38.1	&\textbf{42.1} \\\hline
IBN-Net \cite{pan2018two} &\textcolor{green}{\xmark} &40.3	&\textbf{45.3}	&35.9	&\textbf{44.0}	&35.6	&\textbf{42.1} \\
DRPC \cite{yue2019domain} &\textcolor{green}{\xmark} &42.5	&\textbf{45.8}	&38.0	&\textbf{44.2}	&38.7	&\textbf{42.6} \\
\bottomrule
\end{tabular}}
\label{tab:complement_study}
\vspace{-4mm}
\end{table}

The parameter $p$ is important which controls the sensitivity of DIFs and DVFs identification/selection in Spectrum Learning. We studied the sensitivity of $p$ by changing it from $0$ to $1$ with a step of $1/6$, where `0’ means that all FCs are DVFs (\ie, randomizing all FCs) and `1’ means that all FCs are DIFs (\ie, no randomization). The task is domain generalizable semantic segmentation with GTA and Cityscapes as the source and target domains. Table~\ref{tab:abla_N} shows experimental results. It can be seen that FSDR are not sensitive to $p$ while $p$ lies between $1/6$ and $5/6$. In addition, the FSDR performance drops clearly while $p$ is around either $0$ or $1$, demonstrating the necessity and importance of FC selection in training generalizable networks.

\renewcommand\arraystretch{1.1}
\begin{table}[t]
\caption{The sensitivity of parameter $p$ affects domain generalization: For the task GTA $\rightarrow$ Cityscapes, the domain generalization performance varies with $p$ as evaluated in mIoU.}
\centering
\begin{footnotesize}
\begin{tabular}{ccccccccc}
\hline
\hline
& \multicolumn{7}{c}{Proportion of preserved FCs}
\\\hline
Method & \multicolumn{1}{c}{1} & \multicolumn{1}{c}{5/6} & \multicolumn{1}{c}{4/6} & \multicolumn{1}{c}{3/6} & \multicolumn{1}{c}{2/6} & \multicolumn{1}{c}{1/6} & \multicolumn{1}{c}{0}
\\\hline

FSDR &33.4 &43.8  &44.3     & 44.8  &44.2 &41.0 &38.3\\
\hline
\end{tabular}
\end{footnotesize}
\label{tab:abla_N}
\end{table}

FSDR is also \textit{generic} and can be easily adapted to other tasks. We evaluated this feature by conducting a preliminary UDG-based object detection test that generalizes from SYNTHIA to Cityscapes, Mapillary, and BDD. The experiment setup is the same as what we adopted in the earlier experiments except the change of tasks. Table \ref{Tab:SYNTHIAdet} compares FSDR with a number of UDA and UDG based object detection methods (using ResNet101 as backbone). We can see that FSDR outperforms both domain adaptation and domain generalization methods consistently, demonstrating its genericity in different tasks. Due to the space limit, we place the training details in the supplementary material.

\begin{table}[t]
\centering
\caption{
FSDR is generic and can work for other tasks like object detection: The detection task is SYNTHIA $\rightarrow$ \{\textcolor{red}{C}ityscapes, \textcolor{red}{M}apillary, and \textcolor{red}{B}DDS\} as evaluated using metric mAP. ``w/ Tgt" labels methods that train models with (\textcolor{red}{\cmark} \ie, domain adaptation) and without ( \textcolor{green}{\xmark} \ie, domain generalization) accessing target-domain data in Cityscapes.}
\vspace{2mm}
\label{Tab:SYNTHIAdet}
\resizebox{\columnwidth}{!} {
\begin{tabular}{c|l||c|ccc|c}

\toprule
\hline
Net. & Method & \multicolumn{1}{c|}{\begin{tabular}[c]{@{}c@{}}w/\\ Tgt\end{tabular}} & C & M & B & Mean \\ \hline

\multicolumn{1}{c|}{\multirow{7}{*}{{\begin{tabular}[c]{@{}c@{}}Res\\ Net\\ 101\end{tabular}}}} 
& Faster-RCNN    &\textcolor{green}{\xmark}  & 24.3 & 20.8  &20.1 &21.7 \\
\multicolumn{1}{l|}{}                        & DA~\cite{chen2018domain}  &\textcolor{red}{\cmark} &30.2 &21.2 &21.8 &24.4 \\
\multicolumn{1}{l|}{}                        & MinEnt~\cite{vu2019advent}       &\textcolor{red}{\cmark}     &30.2     & 21.7 &22.4 &24.7     \\
\multicolumn{1}{l|}{}                        & CBST~\cite{zou2018unsupervised}  &\textcolor{red}{\cmark} &32.7 &23.5 &23.9 &26.7 \\
\multicolumn{1}{l|}{}                        & FDA~\cite{yang2020fda}       &\textcolor{red}{\cmark}     &32.4     &23.3 &23.8 &26.5     \\\cline{2-7}
&IBN-Net~\cite{pan2018two} &\textcolor{green}{\xmark} &30.1 &22.3 &23.1 &25.1 \\
& \textbf{Ours (FSDR)} &\textcolor{green}{\xmark} &\textbf{33.5} &\textbf{24.9} & \textbf{25.2} &\textbf{27.8} \\\hline
\end{tabular}}
\end{table}

Due to the space limit, we provide more visualization of FSDR randomized images and qualitative segmentation and detection samples (including their comparisons with state-of-the-art method) in the appendix.

\section{Conclusion}
This paper presents a frequency space domain randomization (FSDR) technique that randomizes images in frequency space by identifying and randomizing domain-variant FCs (DVFs) while keeping domain-invariant FCs (DIFs) unchanged. The proposed FSDR has two unique features: 1) it decomposes images into DIFs and DVFs which allows explicit access and manipulation of them and better control in image randomization; 2) it has minimal effects on image semantics and domain-invariant features. Specifically, we designed spectrum analysis based FSDR (FSDR-SA) and spectrum learning based FSDR (FSDR-SL) both of which can identify DIFs and DVFs effectively. FSDR achieves superior segmentation performance and can be easily incorporated into state-of-the-art domain adaptation and generalization networks with consistent improvement in domain generalization.

\section*{Acknowledgement}
This research was conducted in collaboration with Singapore Telecommunications Limited and supported/partially supported (delete as appropriate) by the Singapore Government through the Industry Alignment Fund - Industry Collaboration Projects Grant.

{\small
\bibliographystyle{ieee_fullname}
\bibliography{egbib}
}
\end{document}